\pdfoutput=1
\PassOptionsToPackage{table}{xcolor}
\documentclass[11pt]{article}

\usepackage{EMNLP2023}

\usepackage{times}
\usepackage{latexsym}
\usepackage{amsmath}
\usepackage{booktabs}
\usepackage{graphicx}
\usepackage{enumitem}
\usepackage[T1]{fontenc}

\usepackage[utf8]{inputenc}
\usepackage{microtype}
\usepackage{colortbl}
\usepackage{float}
\usepackage{algorithm}
\usepackage{algpseudocode}
\usepackage{xspace}
\usepackage{arydshln}
\usepackage{makecell}
\usepackage{amsmath}
\usepackage{mathrsfs}
\usepackage{subfig}
\usepackage{hyperref}
\usepackage{longtable}
\usepackage{inconsolata}
\usepackage{listings}
\usepackage{geometry}
\usepackage{multirow}
\usepackage{booktabs}

\usepackage{array}
\usepackage{adjustbox}
\usepackage{caption}
\usepackage{tabularx}

\definecolor{LightYellow}{rgb}{1.0, 1.0, 0.96}
\definecolor{SkyBlue}{rgb}{0.98, 0.98, 1.0}
\definecolor{LightGreen}{rgb}{0.88, 1.0, 0.88}
\definecolor{LightPink}{rgb}{1.0, 0.98, 0.98}
\definecolor{LightCyan}{rgb}{0.95, 1.0, 1.0}
\definecolor{DarkYellow}{RGB}{255,215,0}
\definecolor{DarkGreen}{RGB}{0,100,0}
\definecolor{DarkBlue}{RGB}{0,0,139}
\definecolor{Brown}{RGB}{165,42,42}

\newcolumntype{y}{>{\columncolor{LightYellow}}c}
\newcolumntype{b}{>{\columncolor{SkyBlue}}c}
\newcolumntype{g}{>{\columncolor{LightGreen}}c}
\newcolumntype{k}{>{\columncolor{LightPink}}c}
\newcolumntype{t}{>{\columncolor{LightCyan}}c}
\newcolumntype{P}[1]{>{\centering\arraybackslash}p{#1}}

\usepackage{graphicx}
\usepackage{epstopdf}

\newcommand{\eg}{\textit{e.g.}\xspace}

\newcommand{\model}{\textsc{LLM}\xspace}

\newcommand{\modelone}{$\textsc{Model}_\textit{per}$\xspace}
\newcommand{\modeltwo}{$\textsc{Model}_\textit{nav}$\xspace}

\newcommand{\llm}{\text{LLM}\xspace}
\newcommand{\llms}{\text{LLMs}\xspace}
\newcommand{\probeloc}{$Probe_\textit{loc}$\xspace}
\newcommand{\probegeo}{$Probe_\textit{geo}$\xspace}

\usepackage{footmisc}

\makeatletter
\def\@fnsymbol#1{\ensuremath{\ifcase#1\or \dagger\or \ddagger\or \mathsection\or \mathparagraph\or \#\or **\or \dagger\dagger\or \ddagger\ddagger \else\@ctrerr\fi}}
\makeatother

\title{Can LLMs Learn to Map the World from Local Descriptions?}

\author{
    Sirui Xia\textsuperscript{\rm$\spadesuit$}, Aili Chen\textsuperscript{\rm$\spadesuit$}, Xintao Wang\textsuperscript{\rm$\spadesuit$}, Tinghui Zhu\textsuperscript{\rm$\spadesuit$}, Yikai Zhang\textsuperscript{\rm$\spadesuit$} \\
    \textbf{Jiangjie Chen\textsuperscript{\rm$\heartsuit$}, Yanghua Xiao\textsuperscript{\rm$\spadesuit$}\thanks{Corresponding authors.}} \\ 
    \textsuperscript{\rm$\spadesuit$}Shanghai Key Laboratory of Data Science, School of Computer Science, Fudan University \\
    \textsuperscript{\rm$\heartsuit$}ByteDance Seed \\
    \texttt{\{srxia24,xtwang21,thzhu22,ykzhang22\}@m.fudan.edu.cn} \quad \\
    \texttt{\{alchen20,shawyh\}@fudan.edu.cn} \quad \texttt{jiangjiec@bytedance.com}  \quad
}

\begin{document}
\maketitle
\begin{abstract}
Recent advances in Large Language Models (LLMs) have demonstrated strong capabilities in tasks such as code and mathematics. However, their potential to internalize structured spatial knowledge remains underexplored. This study investigates whether LLMs, grounded in locally relative human observations, can construct coherent global spatial cognition by integrating fragmented relational descriptions.
We focus on two core aspects of spatial cognition: spatial perception, where models infer consistent global layouts from local positional relationships, and spatial navigation, where models learn road connectivity from trajectory data and plan optimal paths between unconnected locations.
Experiments conducted in a simulated urban environment demonstrate that LLMs not only generalize to unseen spatial relationships between points of interest (POIs) but also exhibit latent representations aligned with real-world spatial distributions. Furthermore, LLMs can learn road connectivity from trajectory descriptions, enabling accurate path planning and dynamic spatial awareness during navigation.
\end{abstract}

\section{Introduction}
\begin{figure}[t]
    \centering
    \includegraphics[width=0.48\textwidth]{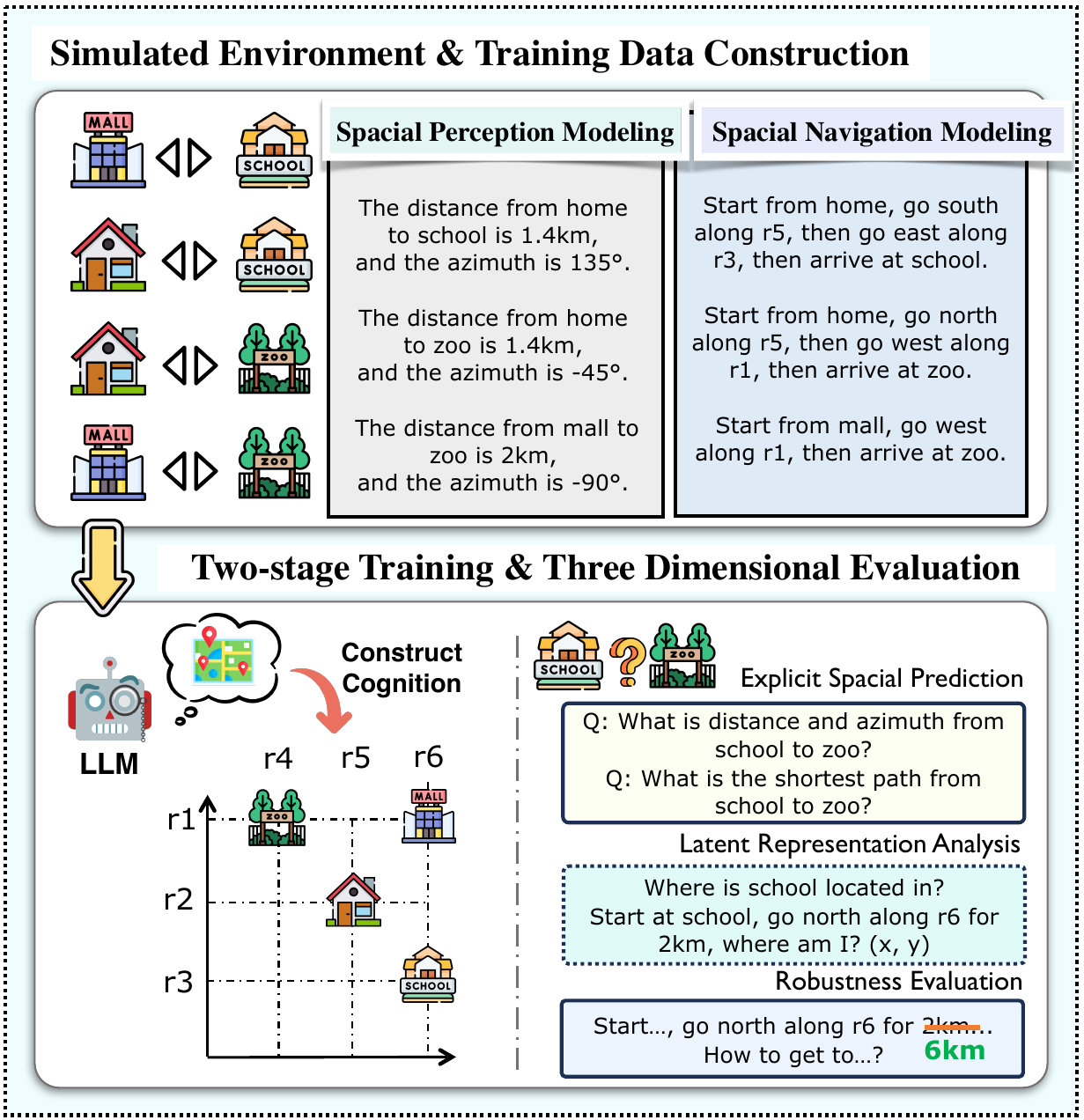}
    \caption{Summary of our research framework. Firsr, we construct a simulated environment and generate training data capturing relative spatial relations and shortest paths. Then, we apply continual pre-training to the \llm and evaluate its spatial cognition through explicit prediction tasks and latent representation analysis.}
    \label{fig:abstract}
\end{figure}

Recent advances in large language models (LLMs) have demonstrated impressive performance across diverse tasks, including code generation, mathematical reasoning, and natural language generation~\citep{chen2021evaluating, shao2024deepseekmath, kojima2022large}.
LLMs are trained on vast amounts of human-generated text~\cite{achiam2023gpt,bai2023qwen}, including structured resources such as Wikipedia and informal unstructured dialogues. 
Since human language inherently relies on local semantic relationships, this enables LLMs to excel at capturing these context-dependent associations.
However, it remains unexplored whether they can implicitly acquire a deep, structured understanding of global information from large amounts of fragmented, localized data—and apply it to reasoning and planning tasks such as spatial reasoning, route optimization, or multi-step inference.

A prime example of a domain requiring structured understanding is \textbf{\textit{spatial cognition}}—the ability to construct coherent mental representations of physical environments.
In human communication, spatial relationships are often conveyed through relational language (\eg, “The library is 100 meters southeast of the park”), which, though concise, encodes rich geometric information such as direction, distance, and topology.
Humans seamlessly integrate such fragmented descriptions into unified mental maps, demonstrating a remarkable capacity to \textit{derive global spatial understanding from localized cues}.
This global cognition supports higher-level spatial reasoning, navigation, and planning.

This reliance on human perspectives raises a fundamental yet underexplored question:
\textbf{\emph{To what extent can LLMs, grounded in locally relative human observations, develop a coherent understanding of global space?}}
This challenge goes beyond processing coordinate data—it requires understanding of spatial language, integrate fragmented descriptions, and build consistent mental maps. 
The model must comprehend geometric relationships (\eg, direction, distance), synthesize incomplete information, and maintain logical coherence across global spatial contexts—all without relying on visual input or explicit coordinates.

To explore this question, we conduct a comprehensive analysis focusing on two core aspects of spatial cognition:
\textbf{1) Spatial Perception}: We investigate whether LLMs can integrate fragmented relational descriptions into a cohesive mental representation of space. This evaluates the model’s ability to reconstruct static spatial understanding solely from linguistic input, without access to explicit coordinate information.
\textbf{2) Spatial Navigation}: We examine whether LLMs can extract topological knowledge from trajectory sequences and apply it to generate optimal paths between previously unconnected locations. This tests the model’s capacity for dynamic spatial reasoning and path planning in the absence of explicit connectivity.

To enable a controlled investigation, we construct a simulated urban environment and introduce a two-stage training and analysis framework guided by two core research questions, which leverages two complementary data modalities:
(1) relational spatial descriptions capturing pairwise distances and directions between points of interest (POIs); and
(2) trajectory descriptions representing shortest paths across the environment.
We analyze whether spatial cognition is formed and how it is expressed through three experimental paradigms:
(1) \textit{Explicit spatial prediction}, assessing task-level prediction;
(2) \textit{Latent representation analysis}, probing geometry in hidden states; and
(3) \textit{Robustness evaluation}, measuring stability under navigational perturbations.
The key findings are as follows:

\begin{itemize}[noitemsep,left=0pt]
\item \textbf{LLMs can construct global spatial cognition from local observations}:
LLMs demonstrate spatial perception by inferring unseen POI relationships, and spatial navigation by planning optimal paths between unconnected locations—revealing coherent global understanding emerging from fragmented linguistic input.
\item \textbf{LLMs can develop structured, implicit spatial representations}:
LLMs encode absolute spatial coordinates within their latent space, aligned with real-world geometry, and dynamically track their position during navigation—indicating the emergence of implicit spatial abstraction without explicit coordinates.
\item \textbf{LLM's spatial navigation remains fragile under perturbation}:
LLMs exhibit limited robustness to path perturbations, with their recovery ability dependent on the distribution of training data, suggesting that their understanding of road spatial information is limited, lacking a continuous and precise representation.
\end{itemize}

\section{Global Setup}
\label{experimental_setup}

\paragraph{Simulation Environment.}
\label{enviroment}
To facilitate controlled investigation and data collection, we construct a synthetic $100 \times 100$ grid map representing a simplified urban layout. Roads run along horizontal and vertical lines ($x = i$ or $y = j$, for $0 \le i, j \le 100$), with traversal weights $w$ randomly sampled from $[0.8, 1.2]$ to simulate varying traffic conditions—higher weights indicate faster travel.
We randomly place $N_{POI} = 1024$ points of interest (POIs) on the grid, each assigned a unique identifier $p_k$ ($k \in {1, 2, \dots, 1024}$). Each grid unit represents 1 kilometer, with the $x$-axis pointing east and the $y$-axis north.

\paragraph{Problem Formulation.}
To explore whether LLMs can develop spatial cognition from natural language descriptions, we define two research tasks that capture key aspects of spatial cognition:
\textbf{(1) \textit{Global Spatial Perception}} — Can the model build a globally consistent understanding of spatial layouts based on local, relational language descriptions?
\textbf{(2) \textit{Dynamic Spatial Planning and Navigation}} — Can the model infer the structure of an underlying road network from local shortest-path descriptions, and use this knowledge to dynamically plan routes between previously unseen pairs of POIs?

\paragraph{Data.} 
\label{data}
\textbf{(1) The \textit{Relational Spatial Dataset}} is used in the first stage to train the model to infer global spatial structure from local pairwise relations. Each sample computes the Euclidean distance $d(p_i, p_j)$ and azimuth $\alpha(p_i, p_j) \in [-180^\circ, 180^\circ]$ between POIs $(p_i, p_j)$, expressed through templated natural language (\eg, “The distance from $p_i$ to $p_j$ is 2.5 km, and the azimuth is 135 degrees.”). To enhance linguistic diversity, we vary the surface realizations of each template.
\textbf{(2) The \textit{Trajectory Dataset}} is used in the second stage to train dynamic spatial navigation. The road network is modeled as a weighted graph, and shortest paths between POIs are computed using Dijkstra’s algorithm. Each path is translated into multi-step natural language instructions~\cite{1959A} (\eg, ``Start at $p_i$, go east on $r_3$ for 3 km, then north on $r_8$ for 2 km to reach $p_j$''), capturing both directional and topological structure.
These datasets are introduced through continuous pre-training in two stages: first, to build coherent spatial representations from relational cues; and second, to acquire navigation capabilities based on learned connectivity.

\paragraph{Model and Two-Stage CPT Training.}
We adopt a two-stage continual pre-training (CPT) framework to explore how spatial cognition emerges in LLMs. CPT allows the model to progressively learn general language and world knowledge from training data, without being constrained by task-specific objectives.
Our focus is on whether an LLM can construct a globally consistent spatial map from localized relational inputs, demonstrating how spatial understanding can be internalized through CPT.
The two training stages correspond to our datasets: the first uses pairwise relational data to foster global spatial perception; the second uses path-based training to develop spatial navigation abilities.
To ensure both interpretability and computational efficiency, we use \textsc{Qwen2.5-0.5B}\cite{Yang2024Qwen25TR} as our base model. Full training details are provided in Appendix\ref{parameter}.

\paragraph{Analysis Approach Overview.}
To systematically investigate the emergence of spatial cognition in LLMs, we design experiments along three complementary dimensions: \textit{functional ability}, \textit{internal representation}, and \textit{behavioral robustness}.
This framework moves beyond surface-level performance to probe the cognitive structures formed during training. Specifically, we assess whether the model can generate accurate spatial predictions, internalize geometry-consistent representations, and maintain stable behavior under perturbations.

\begin{itemize}[noitemsep,left=0pt]
    \item \textbf{\textit{Explicit spatial prediction}} evaluate the model’s ability to perform spatial perception and navigation by predicting distances, azimuths, or shortest paths between unseen POI pairs.

   \item \textbf{\textit{Latent representation analysis}} 
   analyzes the spatial structure encoded in the model’s latent space.
   We apply probing methods to assess whether these representations exhibit geometry-consistent properties, such as encoding absolute coordinates or tracking positions during navigation.

    \item \textbf{\textit{Robustness evaluation}} tests whether the model can navigate accurately under perturbations, focusing on its ability to recover from trajectory deviations and plan effectively under uncertainty.
\end{itemize}

Together, these experiments progress from functional assessment to structural interpretation and robustness evaluation, offering a comprehensive view of how spatial cognition is encoded, composed, and utilized within LLMs.

\section{Modeling Global Spatial Perception from Pairwise Relational Observations}

\label{task1}
In this section, we investigate the capacity of LLMs to develop a holistic understanding of spatial layout from local, spatial relationships, without access to absolute coordinates. 

\subsection{Results of Explicit Spacial Prediction}
\paragraph{Setting.} We begin by evaluating whether the LLM can predict spatial relationships between unseen POI pairs.
We adopt the \textit{relational spatial dataset} in Section~\ref{data}, and evaluate the model's performance under different train-test split ratios. 
We primarily use an 80:20 split, while also testing 60:40 and 40:60. To avoid data leakage, reciprocal POI pairs (e.g., $p_i \rightarrow p_j$ and $p_j \rightarrow p_i$) are always assigned to the same subset. 
We denote the trained model as \modelone.

\paragraph{LLMs exhibit generalized spatial perception across unseen POI pairs.}

\vspace{-1em}
\begin{table}[h!]
  \centering
  \small
  \resizebox{0.48\textwidth}{!}{%
  \begin{tabular}{c yy bb}
    \toprule 

    \multirow{2}{*}{\textbf{Split Ratio}} & \multicolumn{2}{c}{\textbf{Distance}} & \multicolumn{2}{c}{\textbf{Azimuth}} \\
    \cmidrule(lr){2-3} \cmidrule(lr){4-5}

    & \multicolumn{1}{c}{\textbf{MRPE (\%)}~$\downarrow$} & \multicolumn{1}{c}{\textbf{R²}~$\uparrow$} & \multicolumn{1}{c}{\textbf{MRPE (\%)}~$\downarrow$} & \multicolumn{1}{c}{\textbf{Spearman}~$\uparrow$} \\ 
    \midrule

    8:2 & 0.11 & 1.00 & 0.79 & 1.00 \\ 
    6:4 & 0.85 & 1.00 & 3.67 & 0.98 \\ 
    4:6 & 2.63 & 0.99 & 5.36 & 0.98 \\

    \bottomrule 
  \end{tabular}
  }
  \caption{
    Prediction performance on distance and azimuth for unseen POI pairs across different train/test splits.
    MRPE is the Mean Relative Percentage Error; $R^2$ and Spearman reflect consistency in distance and azimuth predictions, respectively.
    }
\label{tab:task1_eval}
\end{table}

\begin{table*}[t!]
  \centering
  \small
  \begin{tabular}{c yyy bbb kk}
    \toprule 

    \multirow{2}{*}{\textbf{Split Ratio}} & \multicolumn{3}{c}{\textbf{X}} & \multicolumn{3}{c}{\textbf{Y}} & \multicolumn{2}{c}{\textbf{Euclidean Distance}} \\
    \cmidrule(lr){2-4} \cmidrule(lr){5-7} \cmidrule(lr){8-9}

     & \multicolumn{1}{c}{\textbf{MSE}~$\downarrow$} & \multicolumn{1}{c}{\textbf{MAE}~$\downarrow$} & \multicolumn{1}{c}{\textbf{R²}~$\uparrow$} & \multicolumn{1}{c}{\textbf{MSE}~$\downarrow$} & \multicolumn{1}{c}{\textbf{MAE}~$\downarrow$} & \multicolumn{1}{c}{\textbf{R²}~$\uparrow$} & \multicolumn{1}{c}{\textbf{Mean}~$\downarrow$} & \multicolumn{1}{c}{\textbf{Std.}~$\downarrow$} \\ 
    \midrule

    Base & 887.76 & 25.99 & -0.01 & 878.72 & 25.10 & -0.10 & 39.19 & 15.18 \\ 
    8:2 & 1.16 & 0.78 & 1.00 & 0.91 & 0.71 & 1.00 & 1.18 & 0.82 \\
    6:4 & 1.30 & 0.76 & 1.00 & 1.55 & 0.82 & 1.00 & 1.26 & 1.12 \\
    4:6 & 2.60 & 1.24 & 1.00 & 3.86 & 1.45 & 1.00 & 2.13 & 1.39 \\

    \bottomrule 
  \end{tabular}
  \caption{Performance of the MLP probe in predicting the absolute coordinates of POIs from the LLM’s last hidden states. Base refers to the untrained LLM. \textbf{X/Y Coordinate Accuracy}: the accuracy of the predicted $x$ and $y$ coordinates using MSE (Mean Squared Error), MAE (Mean Absolute Error) and $R^2$ (Coefficient of Determination). \textbf{Euclidean Distance}: the Euclidean distance between the predicted and true coordinates.}
  \label{tab:task1_rq2_exp1_result}
\end{table*}

\begin{figure}[t!]
    \centering
    \includegraphics[width=0.46\textwidth]{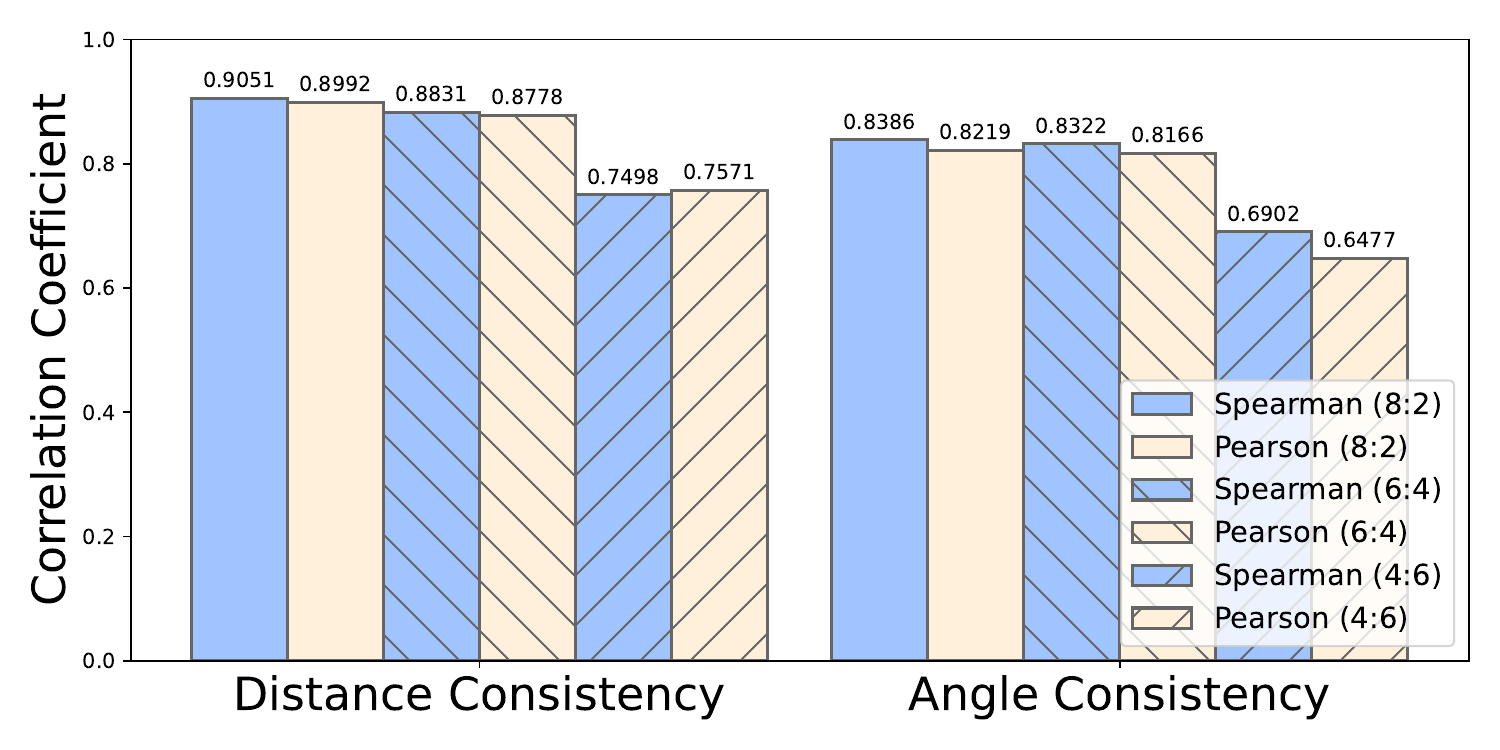}
    \caption{Consistency between POI latent representations and actual spatial locations. Spearman and Pearson correlation coefficients quantify monotonic and linear relationships, respectively.}
    \label{fig:task1_rq2_exp2_result}
\end{figure}

\begin{figure*}[t!]
    \centering
    \includegraphics[width=0.98\textwidth]{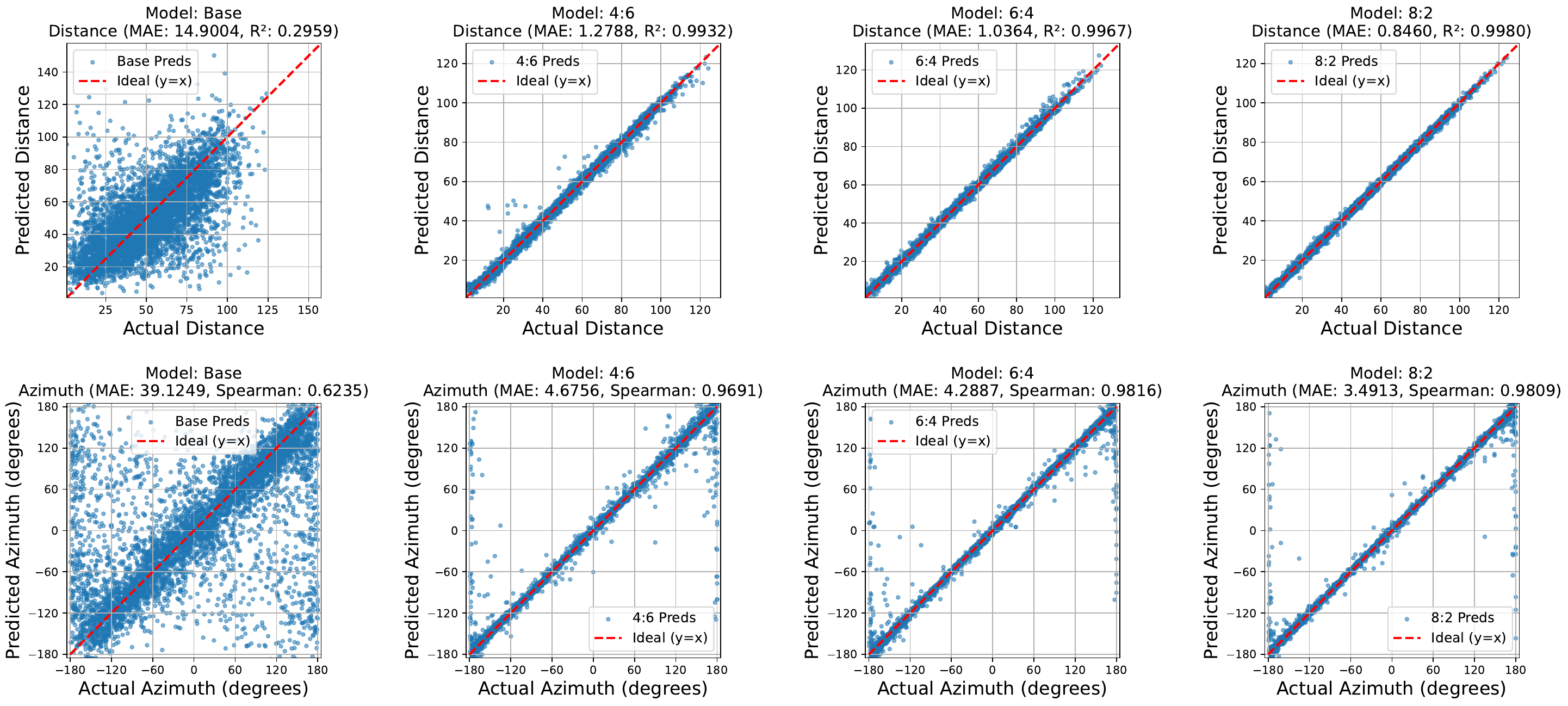}
    \caption{Latent spatial composition evaluation. An MLP predicts distance and azimuth between POI pairs using their concatenated hidden states. We use MAE to measure the deviation between the predicted and true values, and use R² and Spearman correlation to assess the consistency.}
    \label{fig:task1_rq2_exp3_result}
\end{figure*}

As shown in Table~\ref{tab:task1_eval}, \modelone achieves low mean relative percentage errors—0.11\% for distance and 0.71\% for azimuth—demonstrating strong consistency with the ground truth.
This highlights the model’s ability to infer spatial relationships between unseen POI pairs, confirming its success in generalizing spatial perception from local relative relationships.

\paragraph{The strength of generalization is affected by training data scale.}

As the proportion of training data increases, the model’s accuracy in predicting the relative spatial positions of unseen POI pairs improves, with errors decreasing from 2.63\% to 0.11\% across different train/test splits. This trend underscores the critical role of training data scale in enhancing the model’s ability to develop a robust and generalizable global spatial perception.

\subsection{Do LLMs Construct Structured Latent Spatial Representations?}
\label{task1_rq2}

\textbf{Setting.} To investigate whether the model develops spatial perception beyond explicit prediction, we conduct a series of experiments on its latent space. 
These experiments aim to evaluate whether the model encodes spatial coordinate information, how it aligns with physical geometry, and whether spatial relationships can be compositionally inferred.
 
\paragraph{Latent representations encode absolute coordinates.} First, we examine whether the model implicitly encodes absolute POI coordinates by employing an MLP probe (referred to as \probeloc), which investigates the presence of absolute coordinate information in the last hidden state.
Specifically, we encode POI names $p_i$ using \modelone and extract their last hidden states as latent representations. 
These vectors are then fed into \probeloc, a nonlinear regressor that maps them to 2D spatial coordinates $(x, y)$. 
We randomly assign 90\% of the POIs for training and use the remaining 10\% for evaluation. 
The specific MLP configuration is provided in Appendix~\ref{probe_train}.

As shown in Table~\ref{tab:task1_rq2_exp1_result}, predictions from \probeloc yield low Mean Absolute Error, high $R^2$, and small Euclidean deviations, indicating that the last hidden states of \modelone effectively capture absolute coordinate information. 
This suggests that the model not only learns local spatial relations between POIs, but also internalizes a coherent global spatial structure with precise absolute positioning.

\paragraph{Latent spatial layout aligns with physical geometry.}
We further examine the consistency between the last hidden states of POIs and their actual geographic locations. For any three distinct POIs ($p_i$, $p_j$, and $p_k$), we explore two types of spatial consistency:
\textbf{1) \textit{Distance Consistency}}: the correlation between hidden space vector distances ($p_i$-$p_j$, $p_j$-$p_k$, $p_i$-$p_k$) and corresponding Euclidean distances on the map.
\textbf{2) \textit{Angle Consistency}}: the alignment between angles formed by hidden state vectors and those formed by the physical locations.

The results in Figure~\ref{fig:task1_rq2_exp2_result} show a strong alignment between the POIs' spatial layout in latent space and their real-world geography, with consistently high Spearman and Pearson correlations for both distance and angle consistency. 
This suggests that the model’s spatial understanding is internalized in its latent representations, beyond mere prediction accuracy. Unlike probing methods, which train an external model to extract absolute coordinates, this experiment directly examines the latent representations, providing more direct evidence of the model’s structured spatial understanding.

\paragraph{Spatial relations are recoverable via compositional inference over latent representations.}
Building upon these consistent findings, we further investigate whether the latent representations of individual POIs can be compositionally manipulated to infer relative spatial relationships.
For any two POIs, $p_i$ and $p_j$, we extract their last hidden states, concatenate them, and feed the result into an MLP probe (denoted as \probegeo), which is trained to regress a 2D output representing the distance and azimuth between $p_i$ and $p_j$.
We randomly select 100 POIs for evaluation and use the remaining POIs for training. 

Figure~\ref{fig:task1_rq2_exp3_result} shows that \probegeo accurately predicts the distance and azimuth between POIs with low MAE (0.85 \& 3.49) and high $R^2$ (1.00 \& 0.98), indicating that relative spatial relationships between POI pairs can be directly derived by composing their individual POI latent representations. 
This further validates the correctness of the spatial structure captured in the model's latent space and demonstrates the compositionality of its representations, enabling spatial reasoning tasks to be performed directly through the combination of latent vectors.

\begin{table*}[t]
    \small
    \centering
    \scalebox{0.87}{
    \begin{tabular}{l c p{13cm}}
    \toprule
    \textbf{Metric} & \textbf{Full Name} & \textbf{Description} \\
    \midrule
    \textbf{SED} & \textit{Start-End Deviation} & Euclidean distance between the predicted start/end points $(\hat{p}_i, \hat{p}_j)$ and actual POIs $(p_i, p_j)$; composed of Start Point Deviation (SPD) and End Point Deviation (EPD). \\
    
    \textbf{VRP} & \textit{Valid Road Proportion} & Proportion of legal roads selected at each step based on the current position. \\
    
    \textbf{SPA} & \textit{Shortest Path Accuracy} & Fraction of predicted trajectories that exactly match the true shortest path. \\
    
    \textbf{VMR} & \textit{Vector Magnitude Ratio} & Compares straight-line distances between $(p_i, p_j)$ and $(\hat{p}_i, \hat{p}_j)$ to assess distance similarity. \\
    
    \textbf{VCS} & \textit{Vector Cosine Similarity} & Cosine similarity between displacement vectors $p_i \rightarrow p_j$ and $\hat{p}_i \rightarrow \hat{p}_j$, indicating directional consistency. \\
    
    \textbf{FD} & \textit{Fréchet Distance} & Measures geometric similarity between predicted and ground truth trajectories via path point sequences. \\
    
    \textbf{FSA} & \textit{First-Step Accuracy} & Proportion of correct first road selections after applying perturbation to the initial point. \\
    
    \textbf{SA} & \textit{Subsequent Accuracy} & Proportion of correct road selections in all subsequent steps after the first. \\
    
    \textbf{DD} & \textit{Destination Deviation} & Euclidean distance between the final predicted destination and the actual end point. \\
    \bottomrule
    \end{tabular}
    }
    \caption{Evaluation Metrics for Predicted Shortest Paths and Path Perturbations}
    \label{tab:evaluation_metrics}
\end{table*}

\begin{table*}[t!]
  \centering
  \small
  \begin{tabular}{cyyyybbb}
    \toprule
     \multirow{2}{*}{\textbf{Method}} & \multicolumn{4}{c}{\textbf{Accuracy}} & \multicolumn{3}{c}{\textbf{Consistency}} \\
    \cmidrule(lr){2-5} \cmidrule(lr){6-8} 
    
     & \multicolumn{1}{c}{\textbf{SPD}~$\downarrow$} & \multicolumn{1}{c}{\textbf{EPD}~$\downarrow$} & \multicolumn{1}{c}{\textbf{VRP} ($\uparrow$\%)} & \multicolumn{1}{c}{\textbf{SPA} ($\uparrow$\%)} & \multicolumn{1}{c}{\textbf{VMR} ($\uparrow$1.0)} & \multicolumn{1}{c}{\textbf{VCS} ($\uparrow$1.0)} & \multicolumn{1}{c}{\textbf{FD} ($\downarrow$0.0)} \\
    \midrule
    Zero-Exposure (Base) & 49.26 & 49.81 & 87.97 & 0.00 & 0.97 & 0.10 & 58.39 \\
    Zero-Exposure & 5.33 & 10.20 & 94.84 & 0.00 & 0.97 &  0.96 & 13.76 \\
    Bridged Exposure & 0.06 & 0.48 & 96.07 & 83.63 & 1.00 & 1.00 & 0.91 \\
    \bottomrule
  \end{tabular}
  \caption{Performance of different training settings on shortest path prediction between POIs in $P_{\text{heldout}}$. (Base) denotes the model trained on the base model.}
  \label{tab:task2_rq1_result}
\end{table*}

\section{Modeling Spatial Navigation from Local Trajectories}
\label{task2}
We investigate the ability of LLMs to learn road connectivity and spatial navigation capabilities from local trajectory data. The custom evaluation metrics defined in this section are shown in Table~\ref{tab:evaluation_metrics}.

\subsection{Results of Explicit Spacial Prediction}
\paragraph{Setting.} To facilitate generalization analysis, we hold out a subset of 200 POIs (denoted as $P_{\text{heldout}}$), which selectively participate in shortest-path training. 
For the remaining POIs (denoted as $P_{\text{main}}$), we generate shortest-path trajectories for all valid point pairs, and use 80\% of these pairs for training. We denote the trained model as \modeltwo.

\paragraph{Models generalize shortest-path planning to unseen POI pairs by learning from localized trajectories.}
To evaluate model performance under the partially observable condition where all POIs appear as either origins or destinations (but not both) in the training data, we incorporate $P_{\text{heldout}}$ by adding trajectories between $P_{\text{heldout}}$ and $P_{\text{main}}$ POIs, while paths between $P_{\text{heldout}}$ POIs remain unseen (denoted as \textbf{Bridged Exposure} setting).

Table~\ref{tab:task2_rq1_result} shows that \modeltwo excels in shortest-path prediction, with an exact match accuracy of 83.63\% and small start/end deviations (0.06 and 0.48, respectively).
This suggests that the model effectively generalizes road connectivity patterns, not just memorizing seen trajectories, but also performing well on unseen POI pairs.

\paragraph{Models exhibit an emerging ability to compose spatial layout understanding and road network topology for navigation in unseen regions.} 

To further investigate whether the model can leverage the spatial layout understanding established in Section~\ref{task1} to perform shortest-path navigation in unseen regions, we ensure that the POI set $P_{\text{heldout}}$ does not participate in the training data (denoted as \textbf{No-Exposure} setting, these unseen POIs represent unseen regions). 
We then compare the performance between: (1) Perception-\modeltwo - the model trained on \modelone (with spatial layout understanding), and (2) Base-\modeltwo - the model trained on the base model (as baseline).

The results in Table~\ref{tab:task2_rq1_result} reveal that \modeltwo, while trained on \modelone without direct exposure to $P_{\text{heldout}}$ POIs during shortest-path training, performs better than the baseline.
The model shows improvements in both Start-End Deviation (SPD, 49.26~$\rightarrow$ 5.33) and significant gains in directional (VCS, 0.10~$\rightarrow$ 0.96) and geometric (FD, 58.39~$\rightarrow$ 13.76) consistency metrics compared to the baseline. This suggests that while \modeltwo may not yet fully excel at shortest-path navigation in unseen regions, it demonstrates the ability to combine the understanding of POI spatial layout with the understanding of road network topology.

\subsection{Can LLMs Develop Spatial Perception of POI Positions Based on the Shortest Path Trajectory Data?}
\label{task2_rq2}

\begin{table*}[t!]
  \centering
  \small
  \begin{tabular}{c yyy bbb kk}
    \toprule 

    \multirow{2}{*}{\textbf{Model}} & \multicolumn{3}{c}{\textbf{X}} & \multicolumn{3}{c}{\textbf{Y}} & \multicolumn{2}{c}{\textbf{Euclidean Distance}} \\
    \cmidrule(lr){2-4} \cmidrule(lr){5-7} \cmidrule(lr){8-9}

     & \multicolumn{1}{c}{\textbf{MSE}~$\downarrow$} & \multicolumn{1}{c}{\textbf{MAE}~$\downarrow$} & \multicolumn{1}{c}{\textbf{R²}~$\uparrow$} & \multicolumn{1}{c}{\textbf{MSE}~$\downarrow$} & \multicolumn{1}{c}{\textbf{MAE}~$\downarrow$} & \multicolumn{1}{c}{\textbf{R²}~$\uparrow$} & \multicolumn{1}{c}{\textbf{Mean}~$\downarrow$} & \multicolumn{1}{c}{\textbf{Std.}~$\downarrow$} \\ 
    \midrule

    \rowcolor{gray!30}
    \multicolumn{9}{c}{\textit{Absolute Coordinate Probing}} \\
    Base Model & 887.76 & 25.99 & -0.01 & 878.72 & 25.10 & -0.10 & 39.19 & 15.18 \\
    Perception-\modeltwo & 8.53 & 2.16 & 0.99 & 10.21 & 2.40 & 0.99 & 3.54 & 2.49 \\
    Base-\modeltwo & 100.75 & 7.08 & 0.89 & 85.52 & 7.13 & 0.89 & 11.29 & 7.67 \\

    \rowcolor{gray!30}
    \multicolumn{9}{c}{\textit{Step-wise Coordinates Probing}} \\
    Base Model & 713.44 & 19.76 & 0.05 & 621.05 & 18.39 & 0.17 & 30.39 & 20.30 \\
    Perception-\modeltwo & 6.51 & 1.84 & 0.99 & 6.96 & 1.94 & 0.99 & 3.01 & 2.10 \\
    Base-\modeltwo & 22.60 & 2.89 & 0.97 & 21.98 & 2.90 & 0.97 & 4.72 & 4.71 \\

    \bottomrule
  \end{tabular}
  \caption{Performance of the MLP probe in predicting the absolute coordinates of POIs and dynamic position coordinates at each step of the generated navigation path from the model’s last hidden states.}
  \label{tab:task2_rq2_exp2_result}
\end{table*}

\paragraph{Setting.} We next examine whether the model retains spatial perception of POI locations. 
To this end, we compare models trained under the \textbf{Bridged Exposure} setting on \modelone and the base model (denoted as Perception-\modeltwo and Base-\modeltwo).
The untrained base model is also included for comparison.

\paragraph{The model still demonstrates an understanding of the spatial layout of POIs in its latent representations.}

To assess whether the model's latent space still encode absolute coordinate information, we apply the same probing strategy as in Section~\ref{task1_rq2}. As shown in Table~\ref{tab:task2_rq2_exp2_result}, although the spatial perception learned by Base-\modeltwo is less precise than that of Perception-\modeltwo, the model trained solely on shortest-path trajectories shows significant improvements across all evaluation metrics compared to the base model (\eg, X-MAE: 25.99~$\rightarrow$ 7.08, X-R²: -0.01~$\rightarrow$ 0.89).
This demonstrates that even when trained solely on shortest-path trajectories, the model's latent space can encode a certain degree of  absolute coordinate information, highlighting the effectiveness of such data in fostering deeper spatial perception.

\paragraph{The model can dynamically recognize its current position during the navigation process.}

We evaluate the model's ability to encode absolute coordinates at each step of a predicted path using the same probing setup as in previous experiment. 
This allows us to assess whether the model can dynamically track spatial positions as the path unfolds.
To do so, we segment each predicted path into discrete navigation steps (\eg, “go east along $r_1$ for 4km”). 
At each step, we extract the model’s last hidden state from the full input sequence up to that point. The true coordinate of the current location is used as supervision for probe training.
For evaluation, we randomly select 200 POIs as held-out points and use the remaining POIs to construct the training set. We sample 20,000 training trajectories using only the training POIs as both start and end points, and 1,000 evaluation trajectories where the endpoints are drawn from the held-out POIs.

As shown in Table~\ref{tab:task2_rq2_exp2_result}, at each step of the model's navigation, the absolute coordinate position can be clearly extracted from its hidden state (\eg, X-R² 0.05~$\rightarrow$ 0.97). This demonstrates the model's ability to encode and dynamically update its current position at each navigation step, indicating its capacity for dynamic spatial location cognition.

\subsection{Are LLMs Robust to Path Perturbations When Navigating to a Destination?}

\vspace{-1em}
\begin{table}[h!]
  \centering
  \small
  \begin{tabular}{cybk}
    \toprule
    \rowcolor{white}
    \textbf{Type} & \textbf{FSA (\%)} & \textbf{SA (\%)} & \textbf{DD (km)} \\
    \midrule
    No Pert. & 100.00 & 100.00 & 0.00 \\
    Road Pert. & 11.85 & 62.70 & 26.99 \\
    Distance Pert. & 58.79 & 77.71 & 20.24 \\
    Direction Pert. & 43.61 & 74.87 & 56.08 \\
    \bottomrule
  \end{tabular}
  \caption{Navigation performance under various perturbation strategies applied at critical path steps.}
  \label{tab:task2_rq3_exp1_result}
\end{table}

\begin{figure*}[t!]
    \centering
    \begin{minipage}{0.48\textwidth}
        \centering
        \includegraphics[width=\linewidth]{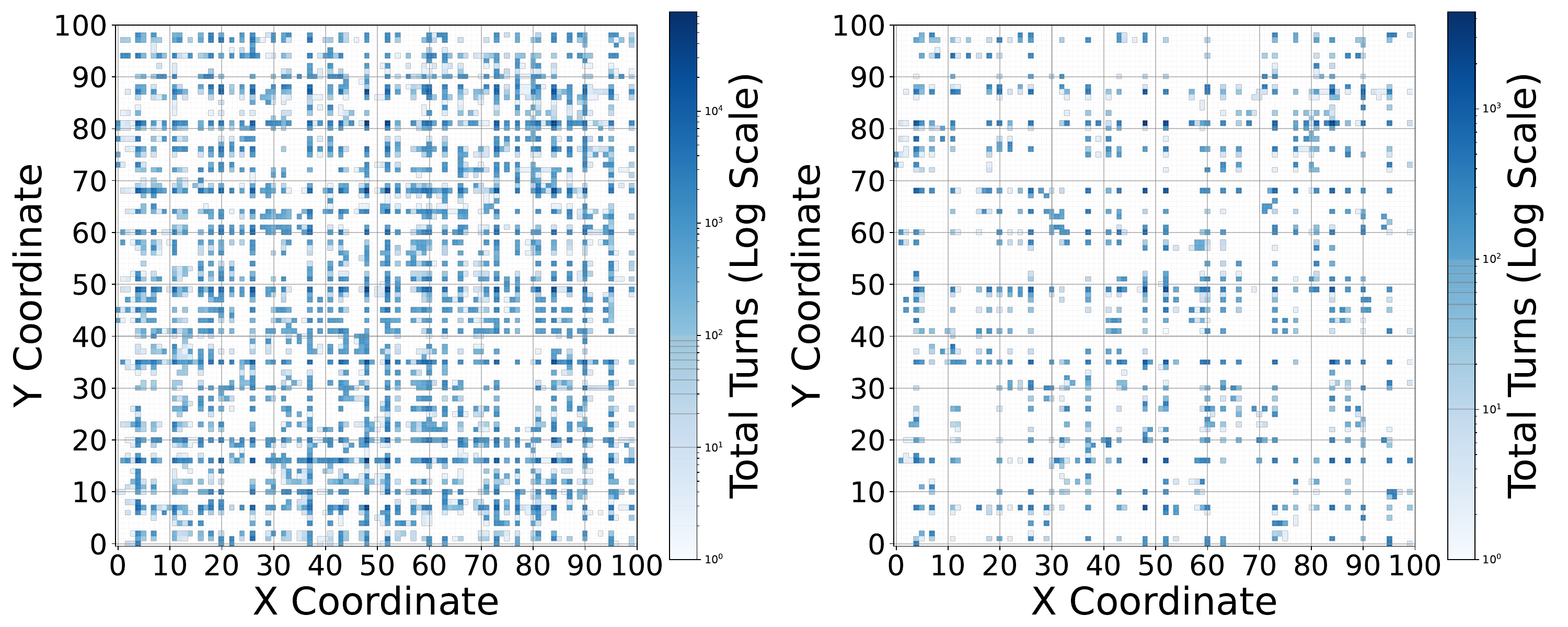}
        \caption{Heatmap of turning point frequencies. The left side shows the training data statistics, while the right side shows the test data statistics.}
        \label{fig:heat_intersection}
    \end{minipage}\hfill
    \begin{minipage}{0.46\textwidth}
        \centering
        \includegraphics[width=\linewidth]{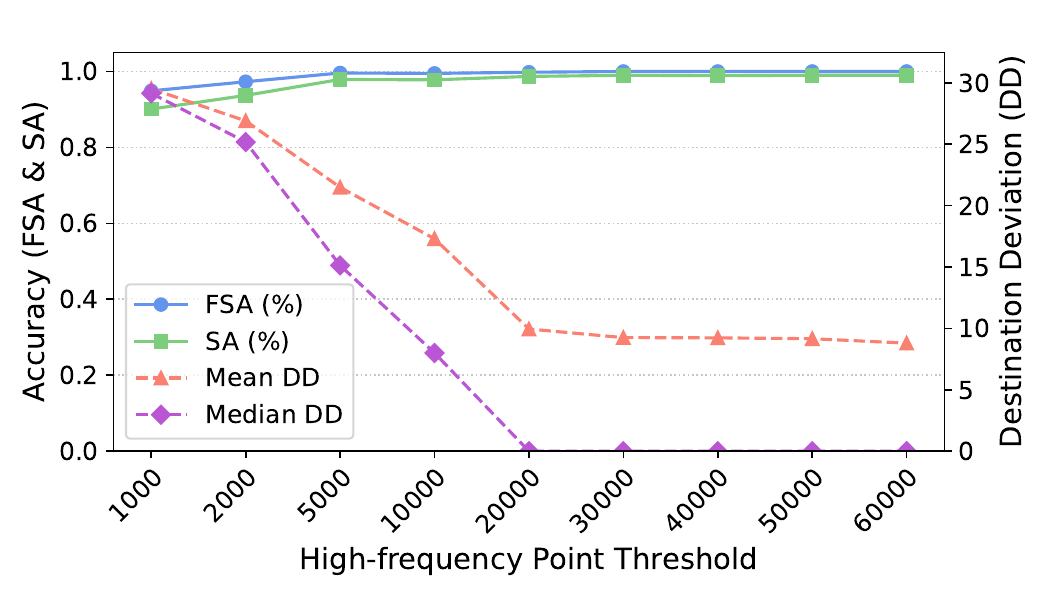}
        \caption{The model's performance under different frequency thresholds.}
        \label{fig:task2_rq3_exp2_result}
    \end{minipage}
\end{figure*}

\paragraph{Setting.} To assess the robustness of the model to trajectory perturbations, we introduce controlled deviations during path prediction to simulate realistic detours, and evaluate whether the model can still reach the intended destination.
These experiments are based on the Perception-\modeltwo defined in Section~\ref{task2_rq2}.
We define $p_{\text{perturb}}$ as the perturbation point and $p_{\text{target}}$ as the immediate location reached after the deviation. Based on this, we design several perturbation strategies.
We identify the step in the predicted trajectory corresponding to the road segment with the highest traversal speed and designate it as the critical step, denoted as $s_{\text{critical}}$.

\paragraph{The model exhibits poor robustness against random disturbances.}

We apply the following types of random perturbations to $s_{\text{critical}}$:
\textbf{1) Road Perturbation}: Replace the original road name in $s_{\text{critical}}$ with a different road (the direction should be modified accordingly);
\textbf{2) Distance Perturbation}: Randomly adjust the distance at $s_{\text{critical}}$, ensuring that it does not exceed the remaining distance to the destination.
\textbf{3) Direction Perturbation}: Invert the heading direction in $s_{\text{critical}}$ (\eg, “east”~$\rightarrow$ “west”).

We select 10,000 cases with original correct predictions by the model for evaluation.

The experimental results in Table~\ref{tab:task2_rq3_exp1_result} show that the model performs poorly when confronted with random perturbations, and its robustness varies across different types of disturbances. 
Specifically, in the road perturbation scenario, the model only has an 11.85\% chance of selecting the first valid passable road, indicating that it does not have a precise understanding of its current location, or it lacks clarity on the available roads at its current position.  This suggests that the model's understanding of the road network is not coherent.

\paragraph{The model's robustness to disturbances largely depends on the distribution of the training data.}

The perturbation experiments reveal a distinction between distance and directional disturbances. 
When subjected to distance perturbations, the model stays on the original high-speed road, \eg, roads within $s_{\text{critical}}$. In contrast, directional perturbations often lead the model to deviate onto slower roads. This may be due to the fact that roads within $s_{\text{critical}}$, which are frequently selected in shortest-path training data and feature higher-frequency entry and exit points along with corresponding turning patterns, are more resilient to disturbances. As a result, they appear to be more robust to distance perturbations but more sensitive to directional ones.

To analyze the impact of turning points, we count the frequency of points in the training data where the direction of movement changes. High-frequency turning points generally correspond to transitions between high-speed and regular roads (Figures~\ref{fig:heat_intersection}). 
We control the selection of $p_{\text{perturb}}$ and $p_{\text{target}}$ by ensuring the location frequency exceeds a threshold $\tau$. 
We select 8,464 cases and analyze how the model's performance varies across different $\tau$.

As shown in Figure~\ref{fig:task2_rq3_exp2_result}, the model's performance improves with an increase in the frequency threshold ($\tau$) for selecting $p_{\text{perturb}}$ and $p_{\text{target}}$. This suggests that the model is more robust to perturbations at high-frequency turning points—enabling it to recover more effectively and reorient towards the correct destination.

These results suggest that although the model exhibits a degree of robustness to perturbations, its recovery ability is highly dependent on the frequency of turning points encountered during training. This reliance implies that the model's understanding of the road network is likely fragmented and localized, rather than comprehensive and global.

\section{Related Work}
\paragraph{World Cognition}
Previous studies have demonstrated that LLMs can encode real-world geospatial~\cite{lietard2021language,treutlein2024connecting} information and temporal~\cite{gurneelanguage} information within their internal representations. 
However, most of these studies use pretrained LLMs in non-anonymized experiments and have not fully explored the source of these capabilities.
Concurrently, many works have focused on the ability of \llms to learn and internalize rules of the physical world or form a “world cognition” of specific tasks from sequential data, such as in board games~\cite{nanda-etal-2023-emergent,li2023emergent,hazinehlinear} or simulated navigation~\cite{pmlr-v235-jin24e,martorell2025textspacemappingabstract,vafa2024evaluating}. 
Unlike predicting the next token based on sequential data, our work focuses on whether LLMs can create a global understanding from natural language descriptions of local observations.

\paragraph{Urban Space Reasoning}

Some works focus on evaluating and enhancing the geospatial reasoning capabilities of LLMs~\cite{feng2024citygpt,feng2024citybench,li2024stbench}. 
These studies mainly focus on enhancing LLMs' spatial reasoning through external information or tools, rather than investigating whether LLMs inherently develop implicit spatial cognition.

\paragraph{Spatial Cognition}
Spatial cognition capabilities are essential for LLMs to understand physical environments and perform tasks involving spatial reasoning.
Many works focus on evaluating and enhancing the spatial cognition capabilities of LLMs~\cite{momennejad2023evaluating,ramakrishnan2024does}, particularly in MLLM settings involving spatial memory and path reasoning~\cite{yang2024thinking,wu2024mind,yu2025spatial}. 
Our work examines text-only LLMs' ability to construct global spatial cognition from localized natural language observations, without relying on global information or coordinates.

\section{Conclusion}
Our study shows that LLMs can develop a global spatial understanding by training on local relative positions and shortest-path data. This is evident in their ability to generalize to unseen POI-pair relationships and in the strong alignment between latent representations and real-world geographic structures. These findings suggest that the model can autonomously build structured spatial cognition from unstructured language to support spatial reasoning. However, its limited robustness to navigation disturbances reveals the constraints of its understanding of road network structures.

\section*{Limitations}
Our study reveals that during the training process, the model develops an understanding of the global spatial distribution of POIs through the description of local relative relationships. 
However, when the model explicitly predicts the positional relationships and the shortest-path trajectories between unseen point pairs, how it utilizes this spatial understanding information has not been fully analyzed. Our experiments lack an in-depth analysis of the internal mechanisms behind the model's explicit prediction of the relative positions and shortest-path trajectories.
Furthermore, our training process has caused the model to lose its original general language capabilities. How to balance the model's general abilities with its internal spatial cognition remains an open research question.

\section*{Ethics Statement}
We hereby acknowledge that all authors of this work are aware of the provided ACL Code of Ethics and honor the code of conduct.

\paragraph{Datasets Source} All studies in this work are based on a simulated, synthetically constructed dataset. The generated data is solely for model analysis research and contains no other usable information. To ensure privacy and ethical compliance, the dataset has been anonymized with placeholder names and contains no real-world information. As a result, the risk of sensitive information leakage is effectively eliminated.

\paragraph{AI assistants} AI assistants (ChatGPT) were solely used to improve the grammatical structure of the text.

\bibliography{anthology,custom}
\bibliographystyle{acl_natbib}

\clearpage
\appendix
\section{Notation Table}

\setlength\tabcolsep{3pt}
\begin{table}[h]
\small
  \centering
    \begin{tabular}{P{0.26\columnwidth}|p{0.72\columnwidth}}
    \toprule
    \textbf{} & \textbf{Definition} \\
    \midrule
    
    \rowcolor[rgb]{ .949,  .953,  .961} \multicolumn{2}{c}{\textit{Task Formulation}} \\
    $\mathcal{G}$ & A graph where intersections are nodes, roads remain unchanged, and the average travel speed is used as the edge weight. \\
    $p_i$ & Names of Points of Interest. \\
    $r_i$ & Names of roads. \\
    \modelone & \model trained on data describing the relative positional relationships between POIs. \\
    \modeltwo & \model trained on data describing the shortest path trajectories. \\
    
    \midrule
    \rowcolor[rgb]{ .949,  .953,  .961} \multicolumn{2}{c}{\textit{Metrics}} \\
    \textbf{MSE} & Mean Squared Error, a metric quantifying the average squared difference between predictions and actual values. \\
    \textbf{MRPE} & Mean Relative Percentage Error, a metric quantifying the average relative percentage difference between predictions and actual values. \\
    \textbf{MAE} & Mean Absolute Error, a metric quantifying the average absolute difference between predictions and actual values. \\
    \textbf{RMSE} & Root Mean Squared Error, a metric quantifying the square root of the average squared difference between predictions and actual values. \\
    \textbf{R²} & R-squared, a metric quantifying the proportion of the variance in the dependent variable that is predictable from the independent variable(s). \\
    \textbf{Spearman} & Spearman correlation coefficient, a metric measuring the strength and direction of the monotonic relationship between two variables. \\
    \textbf{Pearson} & Pearson correlation coefficient, a metric measuring the strength and direction of the linear relationship between two variables. \\
    \textbf{FD} & Fréchet Distance, a metric that measures the similarity between two curves by considering the location and ordering of points. \\
    
    \bottomrule
    \end{tabular}
  \caption{
  The notation table. 
  }
  \label{tab:notations}
\end{table}

In Table~\ref{tab:notations}, we list the notations and abbreviations in this paper, together with their definitions. 

\section{Training Parameters}
\label{parameter}

\paragraph{\model Training} For the continual pre-training of the \model, we use 4$\times$A800 80G GPUs with a batch size of 128, a learning rate of 1.0e$^{-4}$, and a warmup ratio of 0.1, training for 10 epochs. 
Additionally, we designate the POI names $P = \{p_i\}_{i=1}^{1024}$ and road names $R = \{r_i\}_{i=1}^{200}$ as special tokens.

For the SFT of the \model, we train on a single A800 80G GPU with a batch size of 256, a learning rate of 3.0e$^{-5}$, a warmup ratio of 0.1, and train for 10 epochs.
We use Llama-Factory as our training framework~\cite{zheng-etal-2024-llamafactory}.

\paragraph{Probe}
\label{probe_train}
We use the MLPRegression model from scikit-learn~\cite{pedregosa2011scikit}. The MLP probe we use consists of two hidden layers, with 128 and 64 neurons, and ReLU activation functions. 
The model is trained using the Adam optimizer with an initial learning rate of 0.001, and L2 regularization (alpha = 0.0001) with adaptive learning rate adjustment. 
The maximum number of training epochs is set to 500, and early stopping is enabled based on validation set performance (patience = 100 epochs), with a validation set proportion of 10\%. The batch size is adjusted automatically during training, and data is shuffled before each epoch to improve generalization.
All models and tools are publicly available for research purposes.

\section{Experimental Details in Modeling Spatial Cognition}
\label{appendix_task1}

\subsection{POI Distribution}
\label{task1_distribution}

\begin{figure}[ht]
    \centering
    \includegraphics[width=0.48\textwidth]{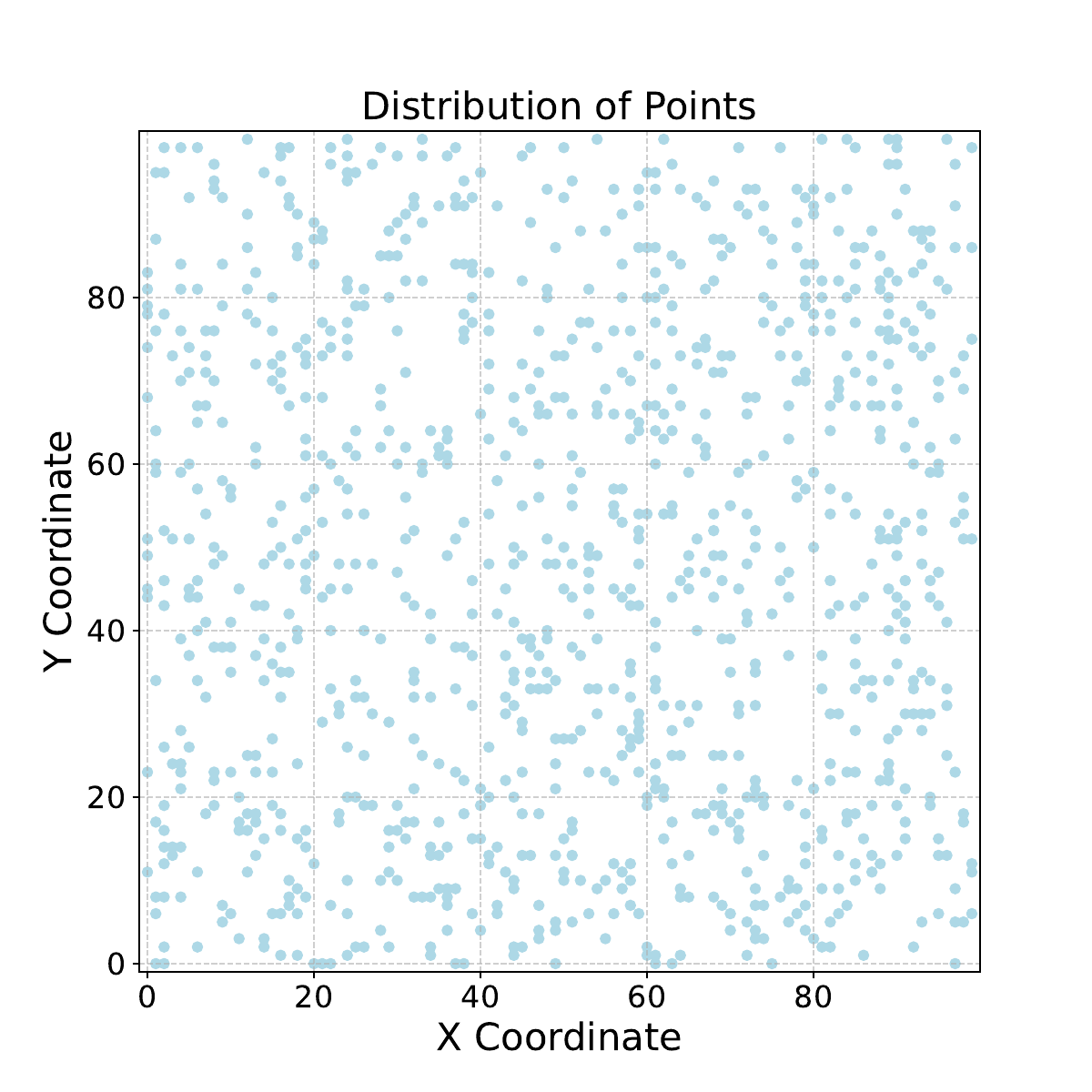}
    \caption{POIs Distribution.}
    \label{fig:task1_poi_distribution}
\end{figure}

The spatial distribution of POI points in Section~\ref{enviroment} is shown in the Figure~\ref{fig:task1_poi_distribution}.

\subsection{Data Format}
\label{task1_data_format}

We provide examples of the data format used for training, as shown in Table~\ref{tab:task1_data_format}.

\begin{table}[h!]
\centering
\small
\renewcommand{\arraystretch}{1.5}
\rowcolors{2}{gray!10}{white}
\begin{tabular}{p{0.48\textwidth}}
\toprule
\textbf{Data Format} \\
\midrule
The distance from $p_i$ to $p_j$ is 1000 meters, with an azimuth of 30 degrees. \\
The distance from $p_i$ to $p_j$ is 1000 meters, and the azimuth from $p_i$ to $p_j$ is 30 degrees. \\
The azimuth from $p_i$ to $p_j$ is 30 degrees, with a distance of 1000 meters. \\
\addlinespace
\textit{Q}: What is the distance from $p_i$ to $p_j$? \newline \textit{A}: 1000 meters. \\
\textit{Q}: What is the azimuth from $p_i$ to $p_j$? \newline \textit{A}: 30 degrees. \\
\textit{Q}: What is the azimuth and distance from $p_i$ to $p_j$? \newline \textit{A}: 30 degrees and 1000 meters. \\
\bottomrule
\end{tabular}
\caption{Different Forms of Training and Evaluation Data for Positional Relationship Description.}
\label{tab:task1_data_format}
\end{table}

\begin{table}[h!]
\centering
\small
\renewcommand{\arraystretch}{1.5}
\rowcolors{2}{gray!10}{white}
\begin{tabular}{p{0.48\textwidth}}
\toprule
\textbf{Data Format} \\
\midrule
Start at $p_i$, then go north on $r_i$ for 2km, then go east on $r_j$ for 10km, and you will arrive at $p_j$. \\
To get from $p_i$ to $p_j$, go along $r_1$ heading north for 2km, then go along $r_2$ heading east for 10km. \\
What is the shortest path from $p_i$ to $p_j$? \newline Answer: First, go north on $r_1$ for 2km, then go east on $r_2$ for 10km. \\
What is the shortest path from $p_i$ to $p_j$? \newline Answer: Go along $r_1$ heading north for 2km, then go along $r_2$ heading east for 10km. \\
\bottomrule
\end{tabular}
\caption{Different Forms of Training and Evaluation Data for Shortest Path Description.}
\label{tab:task2_data_format}
\end{table}


\begin{figure}[htbp]
    \centering
    \includegraphics[width=0.48\textwidth]{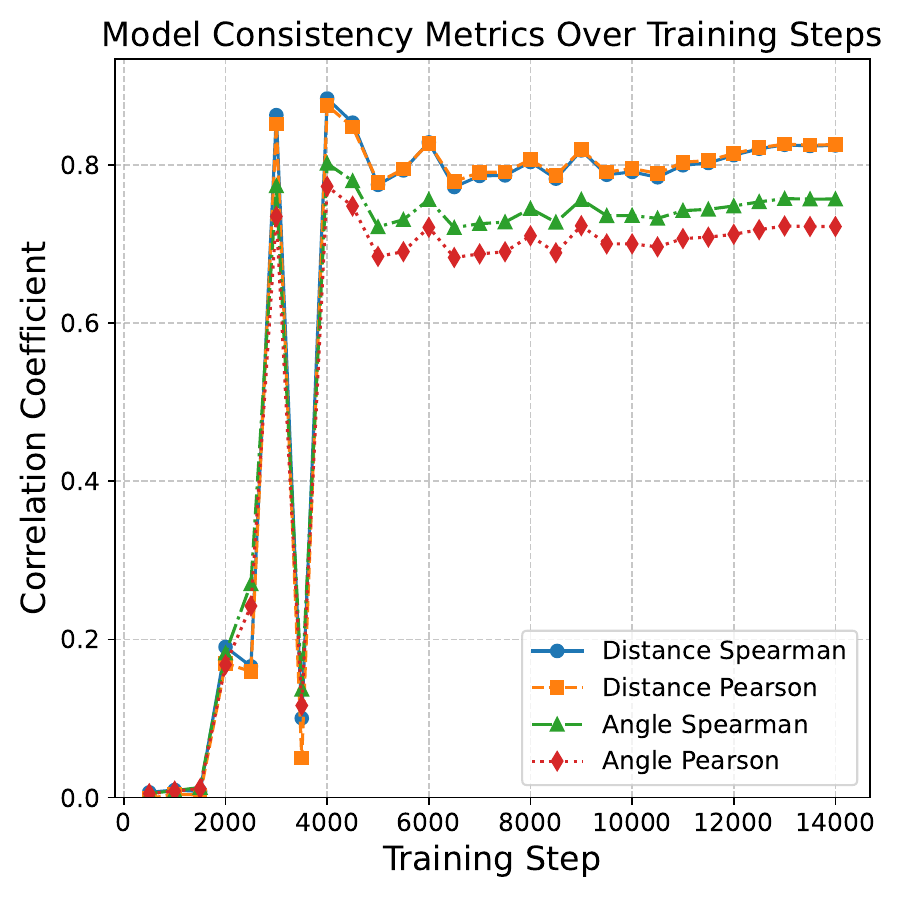}
    \caption{Consistency of POI hidden state vectors with actual spatial locations across training steps.}
    \label{fig:task1_rq2_exp2_trainig}
\end{figure}

\subsection{Additional Experiment}
\paragraph{POI-in-Area Prediction}
\label{task1_rq1_exp2}
We design a simple spatial reasoning task that is inconsistent with the form of the training data in Section~\ref{data}. We train the \modelone through supervised fine-tuning to determine whether a specific POI lies within a given region.
We consider two types of region descriptions: 1) a circular region defined by a central POI and a given radius; 2) a triangular region formed by three POIs. The \model is required to provide a “yes” or “no” answer.

Additionally, we reserve a quarter of the POI points in the Map region, which are not included in the region descriptions of the SFT training data and are only used for evaluation. The remaining POIs are randomly sampled and divided into training and testing sets. We directly use prediction accuracy for evaluation.

\begin{table}[htbp]
  \centering
  \small
  \begin{tabular}{ccc}
    \toprule
   POI Type & Circle (\%) & Triangle (\%) \\
    \midrule
    Included (8:2) & 98.8 & 96.2 \\
    Excluded (8:2) & 97.1 & 97.9 \\
    \hdashline
    Included (6:4) & 98.9 & 97.8 \\
    Excluded (6:4) & 96.1 & 95.8 \\
    \bottomrule
  \end{tabular}
  \caption{Prediction accuracy for POI-in-Area experiment.}
  \label{tab:area_sft}
\end{table}

\section{Experimental Details in Modeling Spatial Navigation}
\label{appendix_task2}

\subsection{Data Format}
\label{task2_data_format}

We provide examples of the data format used for training and evaluation of \modeltwo, as shown in Table~\ref{tab:task2_data_format}.

\begin{table*}[t!]
  \centering
  \small
  \begin{tabular}{cccccccc}
    \toprule
    \multirow{2}{*}{\textbf{Training Strategy}} & \multicolumn{4}{c}{\textbf{Accuracy}} & \multicolumn{3}{c}{\textbf{Consistency}} \\
    \cmidrule(lr){2-5} \cmidrule(lr){6-8} 
    
     & \multicolumn{1}{c}{\textbf{SPD}~$\downarrow$} & \multicolumn{1}{c}{\textbf{EPD}~$\downarrow$} & \multicolumn{1}{c}{\textbf{VRP} ($\uparrow$\%)} & \multicolumn{1}{c}{\textbf{SPA} ($\uparrow$\%)} & \multicolumn{1}{c}{\textbf{VMR} ($\uparrow$1.0)} & \multicolumn{1}{c}{\textbf{VCS} ($\uparrow$1.0)} & \multicolumn{1}{c}{\textbf{FD} ($\downarrow$0.0)} \\
    \midrule
    CPT & 0.06 & 0.48 & 96.07 & 83.63 & 1.00 & 1.00 & 0.91 \\
    SFT & 0.02 & 0.02 & 99.65 & 97.34 & 1.00 & 1.00 & 0.11 \\
    \bottomrule
  \end{tabular}
  \caption{Performance of different training settings on shortest path prediction between POIs in $P_{\text{heldout}}$.}
  \label{tab:train_method_exp5}
\end{table*}

\begin{table*}[t!]
  \centering
  \small
  \begin{tabular}{c yyy bbb kk}
    \toprule

    \multirow{2}{*}{\textbf{Model}} & \multicolumn{3}{c}{\textbf{X}} & \multicolumn{3}{c}{\textbf{Y}} & \multicolumn{2}{c}{\textbf{Euclidean Distance}} \\
    \cmidrule(lr){2-4} \cmidrule(lr){5-7} \cmidrule(lr){8-9}

     & \multicolumn{1}{c}{\textbf{MSE}~$\downarrow$} & \multicolumn{1}{c}{\textbf{MAE}~$\downarrow$} & \multicolumn{1}{c}{\textbf{R²}~$\uparrow$} & \multicolumn{1}{c}{\textbf{MSE}~$\downarrow$} & \multicolumn{1}{c}{\textbf{MAE}~$\downarrow$} & \multicolumn{1}{c}{\textbf{R²}~$\uparrow$} & \multicolumn{1}{c}{\textbf{Mean}~$\downarrow$} & \multicolumn{1}{c}{\textbf{Std.}~$\downarrow$} \\ 
    \midrule

    \rowcolor{gray!30}
    \multicolumn{9}{c}{\textit{Absolute Coordinate Probing}} \\
    Base Model & 887.76 & 25.99 & -0.01 & 878.72 & 25.10 & -0.10 & 39.19 & 15.18 \\
    Cognition-CPT & 8.53 & 2.16 & 0.99 & 10.21 & 2.40 & 0.99 & 3.54 & 2.49 \\
    Base-CPT & 100.75 & 7.08 & 0.89 & 85.52 & 7.13 & 0.89 & 11.29 & 7.67 \\
    Cognition-SFT & 13.05 & 2.89 & 0.98 & 12.88 & 2.76 & 0.99 & 4.39 & 3.84 \\ 
    Base-SFT & 630.21 & 20.83 & 0.25 & 659.85 & 21.14 & 0.25 & 32.55 & 15.19 \\

    \rowcolor{gray!30}
    \multicolumn{9}{c}{\textit{Step-wise Coordinates Probing}} \\
    Base Model & 713.44 & 19.76 & 0.05 & 621.05 & 18.39 & 0.17 & 30.39 & 20.30 \\
    Cognition-CPT & 6.51 & 1.84 & 0.99 & 6.96 & 1.94 & 0.99 & 3.01 & 2.10 \\
    Base-CPT & 22.60 & 2.89 & 0.97 & 21.98 & 2.90 & 0.97 & 4.72 & 4.71 \\
    Cognition-SFT & 11.97 & 2.53 & 0.98 & 12.78 & 2.56 & 0.98 & 4.07 & 3.56 \\
    Base-SFT & 39.01 & 3.91 & 0.95 & 80.64 & 5.13 & 0.89 & 7.50 & 5.21 \\

    \bottomrule
  \end{tabular}
  \caption{Performance of the MLP probe in predicting the absolute coordinates of POIs and dynamic position coordinates at each step of the generated navigation path from the \model’s last hidden states.}
  \label{tab:train_method_exp6}
\end{table*}

\subsection{Metric Calculations}
\label{metric_calculation}

\paragraph{Start-End Deviation (SED)}: evaluates the spatial accuracy of the predicted path description by computing the Euclidean distance between predicted and ground truth coordinates at both the start and end points. The predicted trajectory is reconstructed by simulating the movement along a parsed sequence of road-based navigation steps using map information. The final metric is reported as a tuple: Start Deviation (SD) and End Deviation (ED). Detailed computation logic is provided in Algorithm~\ref{alg:sed_refined}.

\paragraph{Valid Road Proportion (VRP)}: measures the proportion of valid road choices at each step of the predicted path description. The path is parsed into a sequence of steps, and for each step, the algorithm checks if the road and direction are valid according to the map's connectivity and direction rules. The final metric, VRP, is the ratio of valid steps to the total steps in the path description. If no steps are described, the VRP is defined as 0. Detailed computation logic is provided in Algorithm~\ref{alg:vrp}.

\paragraph{Shortest Path Accuracy (SPA)}: measures the proportion of cases where the model-generated trajectory exactly matches the ground truth shortest path. 

\subsection{Case Study}

\begin{figure*}[ht]
    \centering
    \includegraphics[width=0.98\textwidth]{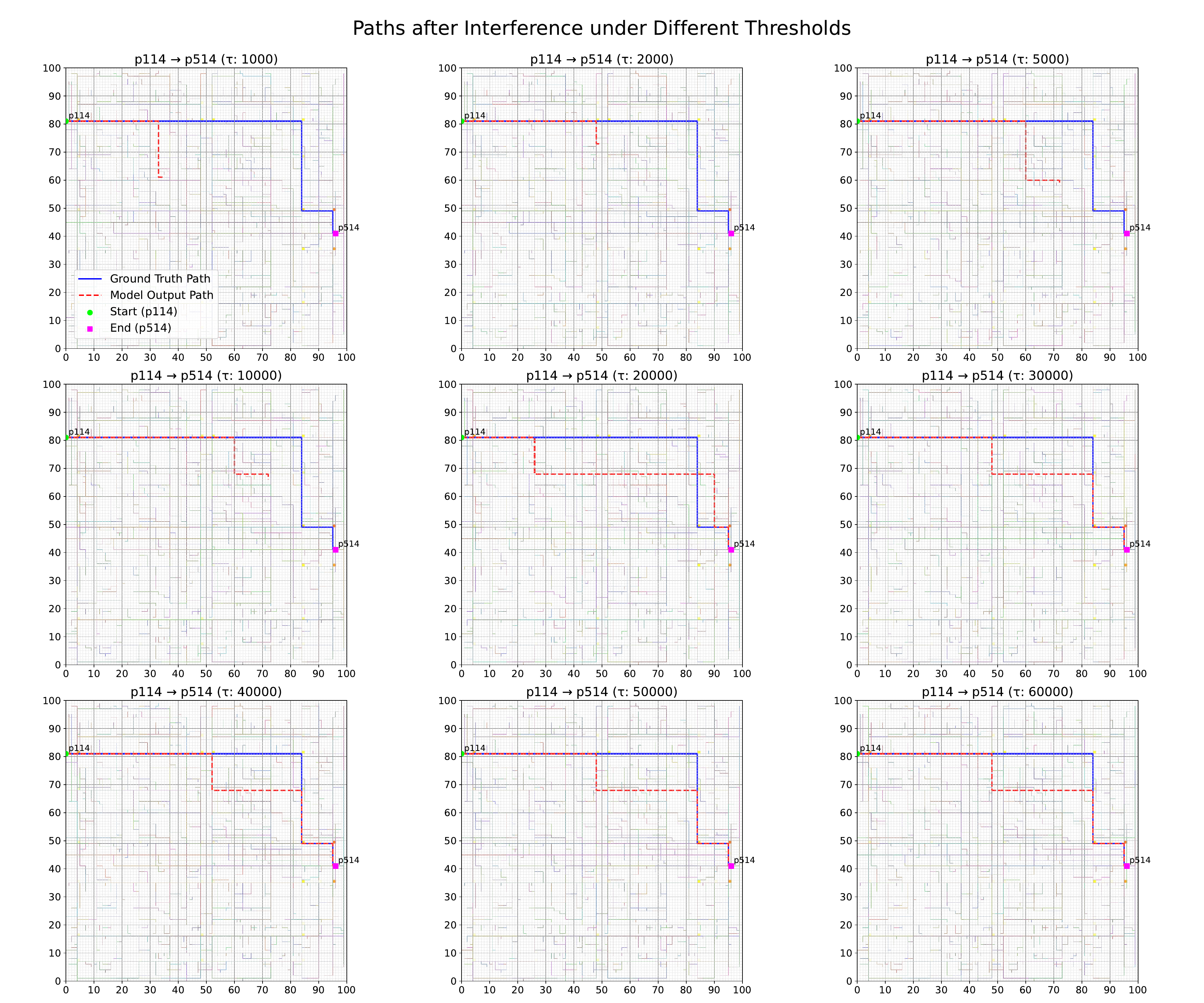}
    \caption{Case study on the model's behavior under interference during navigation at different statistical frequencies.}
    \label{fig:case_study}
\end{figure*}

Figure~\ref{fig:case_study} illustrates the \model's performance in handling intermediate disturbances under different turning point frequencies. As the frequency increases, the \model becomes more robust to disturbances and is able to reach the final destination after the disturbance.

\subsection{Additional Experiment}

\begin{table}[h!]
  \centering
  \resizebox{0.48\textwidth}{!}{
    \begin{tabular}{c y b}
      \toprule
      \rowcolor{white}
      \textbf{Model} & \textbf{Distance \%}~$\downarrow$ & \textbf{Azimuth \%}~$\downarrow$ \\
      \midrule
      Perception-\modeltwo & 3.08 & 5.52 \\
      Base-\modeltwo & 12.03 & 13.84  \\
      \bottomrule
    \end{tabular}
  }
  \caption{Evaluation results for distance and azimuth prediction, evaluated using MRPE.}
  \label{tab:task2_rq2_exp1_result}
\end{table}

\paragraph{The model remains capable of performing explicit spatial relationship prediction.}
To assess whether the model directly trained on path data can still understand the relative positional relationships between POIs, we fine-tune it with supervised training to predict the distance and azimuth between POI pairs. 
We use 200 POIs to construct the test set, while the remaining POIs are used to generate the training data (randomly sample 100,000 cases).

The results in Table~\ref{tab:task2_rq2_exp1_result} show that training the base model on shortest-path trajectories (Base-\modeltwo) allows it to capture the relative spatial relationships between POI pairs, achieving reasonable performance in both distance and azimuth prediction, with MRPE values of 12.03\% and 13.84\%, respectively. 
This suggests that, even without directly relying on local distance and azimuth information between POI pairs, the model is still able to leverage shortest-path trajectories to build a certain level of global spatial perception. 
This also indicates that shortest-path trajectories, as a topologically structured data format, are effective in constructing an understanding of spatial layout.

\begin{algorithm*}[ht]
\caption{$SED$: Start-End Deviation Calculation}\label{alg:sed_refined}
\begin{algorithmic}[1]
\State \textbf{Input}: Ground truth start coordinates $P_{start\_gt}$, Ground truth end coordinates $P_{end\_gt}$, \model-generated textual path description $\mathcal{A}$, Map information $\mathcal{M}_{map}$
\State \textbf{Output}: Start-End Deviation $SED$ \Comment{Euclidean distance between predicted and ground truth points}

\State $\mathcal{S} \gets \text{ParsePathDescription}(\mathcal{A})$ \Comment{Parse $\mathcal{A}$ into sequence of steps $\mathcal{S} = [(r_1, d_1, l_1), \dots, (r_n, d_n, l_n)]$}

\If{$|\mathcal{S}| < 2$}
    \State $P_{start\_pred} \gets P_{start\_gt}$ \Comment{Use ground truth start if path description has fewer than 2 steps}
\Else
    \State Let $(r_1, d_1, l_1) = \mathcal{S}[1]$ \Comment{First step details}
    \State Let $(r_2, d_2, l_2) = \mathcal{S}[2]$ \Comment{Second step details}
    \State $P_{intersect} \gets \text{FindIntersection}(r_1, r_2, \mathcal{M}_{map})$ \Comment{Find intersection of the first two roads (position after the first step)}
    \If{$P_{intersect}$ is valid} \Comment{Check if a valid intersection was found}
        \State $P_{start\_pred} \gets \text{MoveAlongRoad}(P_{intersect}, r_1, \text{Opposite}(d_1), l_1, \mathcal{M}_{map})$ \Comment{Backtrack from intersection to estimate start}
    \Else
        \State $P_{start\_pred} \gets P_{start\_gt}$ \Comment{Fallback to ground truth start if intersection is indeterminate}
    \EndIf
\EndIf

\State $P_{current} \gets P_{start\_pred}$ \Comment{Initialize current position}

\If{$|\mathcal{S}| > 0$} \Comment{Simulate the path if steps exist}
    \For{each step $(r_i, d_i, l_i)$ in $\mathcal{S}$}
        \State $P_{current} \gets \text{MoveAlongRoad}(P_{current}, r_i, d_i, l_i, \mathcal{M}_{map})$ \Comment{Update position}
    \EndFor
\EndIf
\State $P_{end\_pred} \gets P_{current}$ \Comment{The final position is the predicted end position}

\State $SD \gets \text{EuclideanDistance}(P_{start\_pred}, P_{start\_gt})$ \Comment{Calculate Start Deviation}
\State $ED \gets \text{EuclideanDistance}(P_{end\_pred}, P_{end\_gt})$ \Comment{Calculate End Deviation}
\State \Return $(SD, ED)$ \Comment{Return deviations at both start and end points}

\Statex 
\Statex \Comment{Helper Functions:}
\Statex \Comment{- $\text{ParsePathDescription}(\mathcal{A})$: Parses the textual path description $\mathcal{A}$ into a structured list $\mathcal{S}$ of tuples, where each tuple is $(road\_id, direction, length)$.}
\Statex \Comment{- $\text{FindIntersection}(r_a, r_b, \mathcal{M}_{map})$: Returns the geographic coordinates of the intersection between road segment $r_a$ and road segment $r_b$ based on $\mathcal{M}_{map}$. Returns an invalid/null state if no relevant intersection exists.}
\Statex \Comment{- $\text{MoveAlongRoad}(P_{origin}, r, d, l, \mathcal{M}_{map})$: Calculates the coordinates resulting from starting at $P_{origin}$, moving along road $r$ in direction $d$ for distance $l$, according to $\mathcal{M}_{map}$.}
\Statex \Comment{- $\text{Opposite}(d)$: Returns the direction directly opposite to $d$ (e.g., Opposite(North) = South).}
\Statex \Comment{- $\text{EuclideanDistance}(P_1, P_2)$: Computes the L2 norm (straight-line distance) $||P_1 - P_2||_2$.}
\end{algorithmic}
\end{algorithm*}

\begin{algorithm*}[ht]
\caption{$VRP$: Valid Road Proportion Calculation}\label{alg:vrp}
\begin{algorithmic}[1]
\State \textbf{Input}: Ground truth start coordinates $P_{start\_gt}$, \model-generated textual path description $\mathcal{A}$, Map information $\mathcal{M}_{map}$
\State \textbf{Output}: Valid Road Proportion $VRP$ \Comment{Proportion of steps choosing a valid next road}

\State $\mathcal{S} \gets \text{ParsePathDescription}(\mathcal{A})$ \Comment{Parse $\mathcal{A}$ into sequence of steps $\mathcal{S} = [(r_1, d_1, l_1), \dots, (r_n, d_n, l_n)]$}

\If{$|\mathcal{S}| < 2$}
    \State $P_{start\_pred} \gets P_{start\_gt}$ \Comment{Use ground truth start if path description has fewer than 2 steps}
\Else
    \State Let $(r_1, d_1, l_1) = \mathcal{S}[1]$ \Comment{First step details}
    \State Let $(r_2, d_2, l_2) = \mathcal{S}[2]$ \Comment{Second step details}
    \State $P_{intersect} \gets \text{FindIntersection}(r_1, r_2, \mathcal{M}_{map})$ \Comment{Find intersection of the first two roads (position after the first step)}
    \If{$P_{intersect}$ is valid} \Comment{Check if a valid intersection was found}
        \State $P_{start\_pred} \gets \text{MoveAlongRoad}(P_{intersect}, r_1, \text{Opposite}(d_1), l_1, \mathcal{M}_{map})$ \Comment{Backtrack from intersection to estimate start}
    \Else
        \State $P_{start\_pred} \gets P_{start\_gt}$ \Comment{Fallback to ground truth start if intersection is indeterminate}
    \EndIf
\EndIf

\State $P_{current} \gets P_{start\_pred}$ \Comment{Initialize current position}
\State $valid\_steps \gets 0$ \Comment{Initialize counter for valid road choices}
\State $total\_steps \gets |\mathcal{S}|$ \Comment{Total number of steps in the described path}

\If{$total\_steps > 0$} \Comment{Simulate the path if steps exist}
    \For{each step $(r_i, d_i, l_i)$ in $\mathcal{S}$}
        \State $\mathcal{R}_{valid} \gets \text{GetValidNextRoads}(P_{current}, \mathcal{M}_{map})$ \Comment{Get set of valid (road\_name, road\_direct)}
        \If{$(r_i, d_i) \in \mathcal{R}_{valid}$} \Comment{Check if the chosen road and direction are valid options}
            \State $valid\_steps \gets valid\_steps + 1$ \Comment{Increment valid step count}
        \EndIf
        \State $P_{current} \gets \text{MoveAlongRoad}(P_{current}, r_i, d_i, l_i, \mathcal{M}_{map})$ \Comment{Update position}
    \EndFor
\EndIf

\If{$total\_steps == 0$}
    \State $VRP \gets 0$ \Comment{Define VRP as 0 for empty paths}
\Else
    \State $VRP \gets valid\_steps / total\_steps$ \Comment{Calculate the proportion of valid steps}
\EndIf

\State \Return $VRP$

\Statex 
\Statex \Comment{Helper Functions:}
\Statex \Comment{- $\text{ParsePathDescription}(\mathcal{A})$: Parses the textual path description $\mathcal{A}$ into a structured list $\mathcal{S}$ of tuples $(road\_id, direction, length)$.}
\Statex \Comment{- $\text{MoveAlongRoad}(P_{origin}, r, d, l, \mathcal{M}_{map})$: Calculates coordinates after moving from $P_{origin}$ along road $r$ in direction $d$ for distance $l$.}
\Statex \Comment{- $\text{Opposite}(d)$: Returns the direction opposite to $d$.}
\Statex \Comment{- \(\text{GetValidNextRoads}(P_{pos}, \mathcal{M}_{map})\): Returns a set of valid next moves as (road\_id, direction) tuples accessible from position \(P_{pos}\). This considers connectivity and travel direction rules based on map data \(\mathcal{M}_{map}\).}
\Statex \Comment{- $\text{EuclideanDistance}(P_1, P_2)$: Computes the L2 norm $||P_1 - P_2||_2$. (Included for consistency, though not used in VRP calculation itself).}
\end{algorithmic}
\end{algorithm*}

\section{Additional Experiments and Results}
\label{additional_exp}

\subsection{Training Strategy}
\label{train_method}

\begin{table}[htbp]
  \centering
  \small
  \begin{tabular}{c cc cc}
    \toprule

    \multirow{2}{*}{\textbf{Training Strategy}} & \multicolumn{2}{c}{\textbf{Distance}} & \multicolumn{2}{c}{\textbf{Azimuth}} \\
    \cmidrule(lr){2-3} \cmidrule(lr){4-5}

    & MRPE~$\downarrow$ & R²~$\uparrow$ & MRPE~$\downarrow$ & Spearman~$\uparrow$ \\ 
    \midrule

    CPT & 0.11 & 1.00 & 0.79 & 1.00 \\ 
    SFT & 0.003 & 1.00 & 0.025 & 1.00 \\

    \bottomrule
  \end{tabular}
  \caption{The performance of the model's prediction of distance and azimuth for unseen POI pairs under different training strategies.}
  \label{tab:train_method_exp1}
\end{table}

\begin{table*}[htbp]
  \centering
  \small
  \begin{tabular}{c ccc ccc cc}
    \toprule

    \multirow{2}{*}{\textbf{Training Strategy}} & \multicolumn{3}{c}{\textbf{X}} & \multicolumn{3}{c}{\textbf{Y}} & \multicolumn{2}{c}{\textbf{Euclidean Distance}} \\
    \cmidrule(lr){2-4} \cmidrule(lr){5-7} \cmidrule(lr){8-9}

     & MSE~$\downarrow$ & MAE~$\downarrow$ & R²~$\uparrow$ & MSE~$\downarrow$ & MAE~$\downarrow$ & R²~$\uparrow$ & Mean~$\downarrow$ & Std.~$\downarrow$ \\ 
    \midrule

    Base & 887.76 & 25.99 & -0.01 & 878.72 & 25.10 & -0.10 & 39.19 & 15.18 \\ 
    CPT & 1.16 & 0.78 & 1.00 & 0.91 & 0.71 & 1.00 & 1.18 & 0.82 \\
    SFT & 406.66 & 15.41 & 0.46 & 373.35 & 14.23 & 0.53 & 23.10 & 15.69 \\

    \bottomrule 
  \end{tabular}
  \caption{Performance of the MLP probe in predicting the absolute coordinates of POIs from the \model’s last hidden states under different training strategies.}
  \label{tab:train_method_exp2}
\end{table*}

\begin{table}[h]
  \centering
  \small
  \begin{tabular}{c cc cc}
    \toprule 

    \multirow{2}{*}{\textbf{Training Strategy}} & \multicolumn{2}{c}{\textbf{Distance}} & \multicolumn{2}{c}{\textbf{Azimuth}} \\
    \cmidrule(lr){2-3} \cmidrule(lr){4-5}

     & MAE (km) & R² & MAE (°) & Spearman  \\ 
    \midrule

    Base & 14.90 & 0.03 & 39.12 & 0.62 \\ 
    CPT  & 0.85 & 1.00 & 3.49 & 0.98 \\
    SFT & 31.62  & -2.92 & 66.48 & 0.38 \\ 

    \bottomrule
  \end{tabular}
  \caption{Latent spatial composition evaluation. An MLP predicts distance and azimuth between POI pairs using their concatenated hidden states.}
  \label{tab:train_method_exp4}
\end{table}

\begin{figure}[h]
    \centering
    \includegraphics[width=0.48\textwidth]{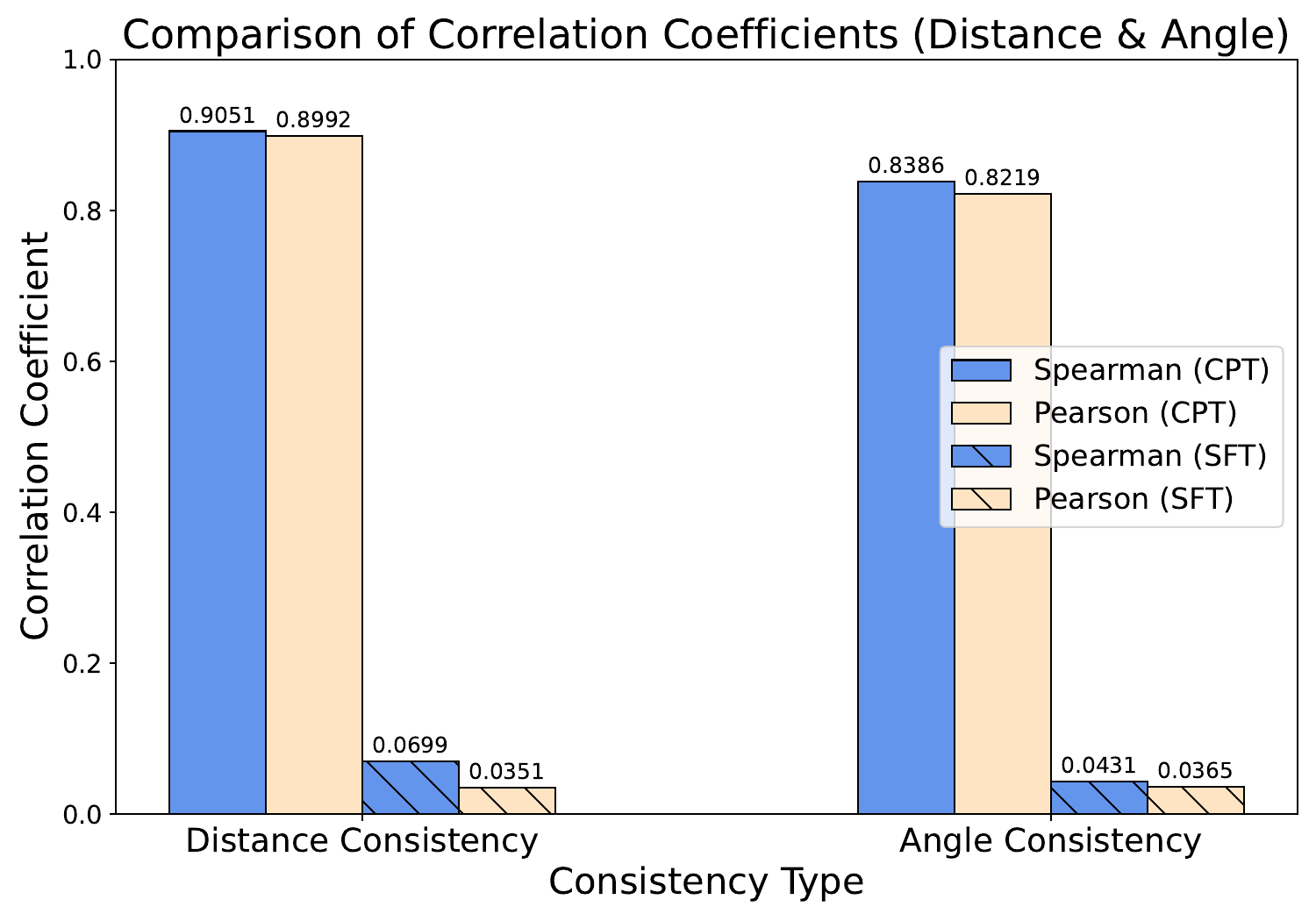}
    \caption{Consistency of POI point last hidden state vector with actual spatial location in terms of distance and angle under different training strategies.}
    \label{fig:train_method_exp3}
\end{figure}

Our primary experiments adopt a continual pre-training approach for \model training. In addition to this, we explore the use of SFT for training \modelone and \modeltwo. 
For training \modelone, we retain the question-answer format data from the original complete dataset and adopt an 80/20 split for training and test sets. For \modeltwo, we follow the \textbf{Bridged Exposure} strategy.
We evaluate whether the \model trained with SFT can perform explicit predictions and construct cognitive representations in the latent space.

We conduct training using 4×A800 80G GPUs, with a batch size of 512 and a learning rate set to 3.0e-5. The \model is trained for 10 epochs.

\paragraph{Spatial Cognition}
The experimental results for evaluating Spatial Cognition are shown in Table~\ref{tab:train_method_exp1}, Table~\ref{tab:train_method_exp2}, Table~\ref{tab:train_method_exp4} and Figure~\ref{fig:train_method_exp3}.

Experimental results show that while SFT-trained \model outperform CPT-trained \model in distance and azimuth prediction accuracy, they exhibit weaker latent spatial cognition, as evidenced by blurred awareness of absolute coordinates in hidden states and poor alignment between latent vector distributions and actual spatial layouts.

This result is expected, as the POI name tokens in the SFT training process do not directly contribute to the loss calculation. Consequently, their embeddings are not explicitly optimized, leading to a lack of structured distribution in the latent space.
This highlights the importance of continual pre-training for fostering deeper internal representations.
At the same time, it suggests that a well-structured latent distribution of individual POIs is not strictly necessary for predicting relative relationships between unseen POI pairs.

\paragraph{Spatial Navigation}
The experimental results for evaluating Spatial Navigation are shown in Table~\ref{tab:train_method_exp5} and Table~\ref{tab:train_method_exp6}.

\begin{figure*}[t!]
    \centering
    \begin{minipage}{0.48\textwidth}
        \centering
        \includegraphics[width=\linewidth]{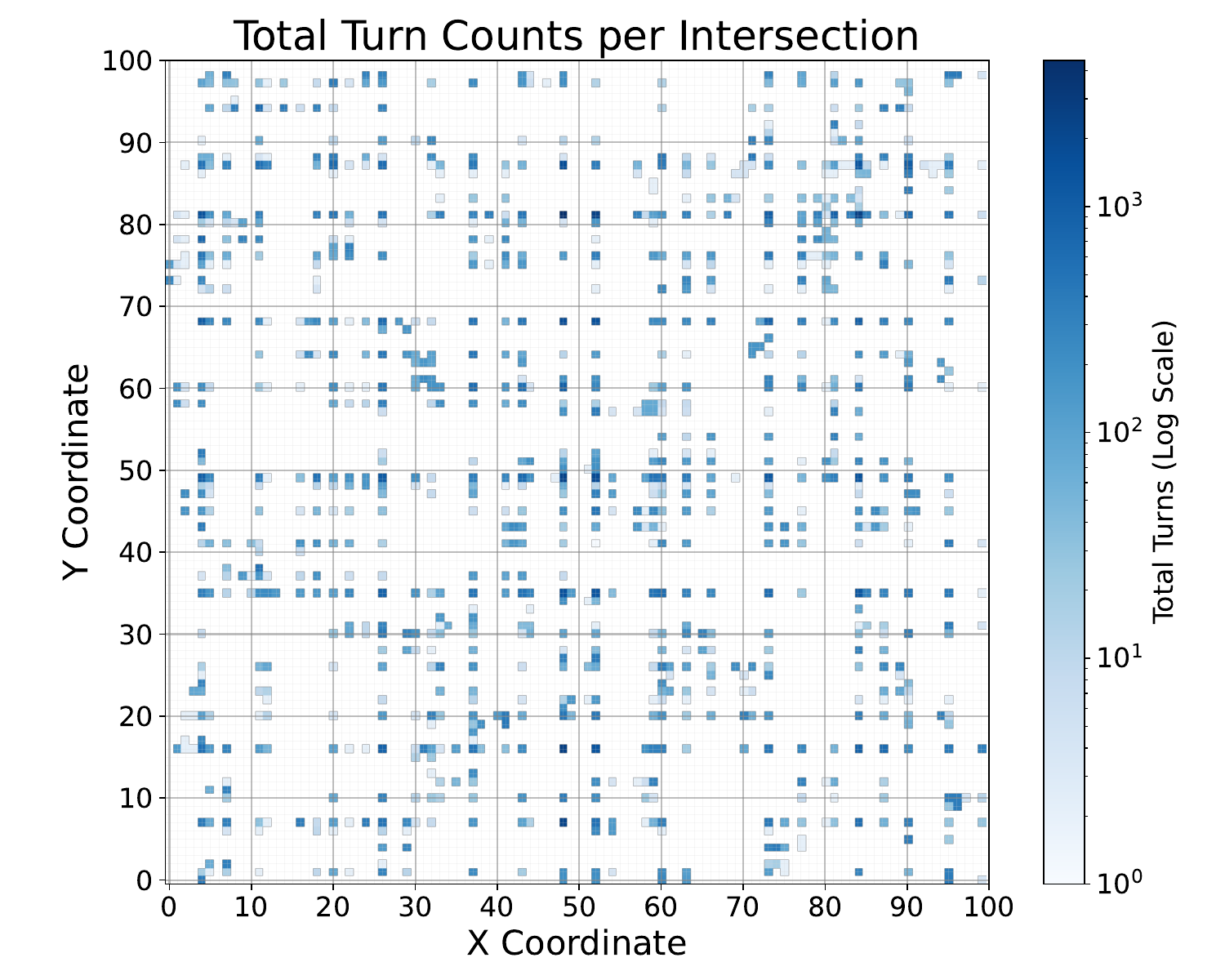}
        \caption{Heatmap of turning point frequencies in the evaluation data.}
        \label{fig:turn_count_eval}
    \end{minipage}\hfill
    \begin{minipage}{0.48\textwidth}
        \centering
        \includegraphics[width=\linewidth]{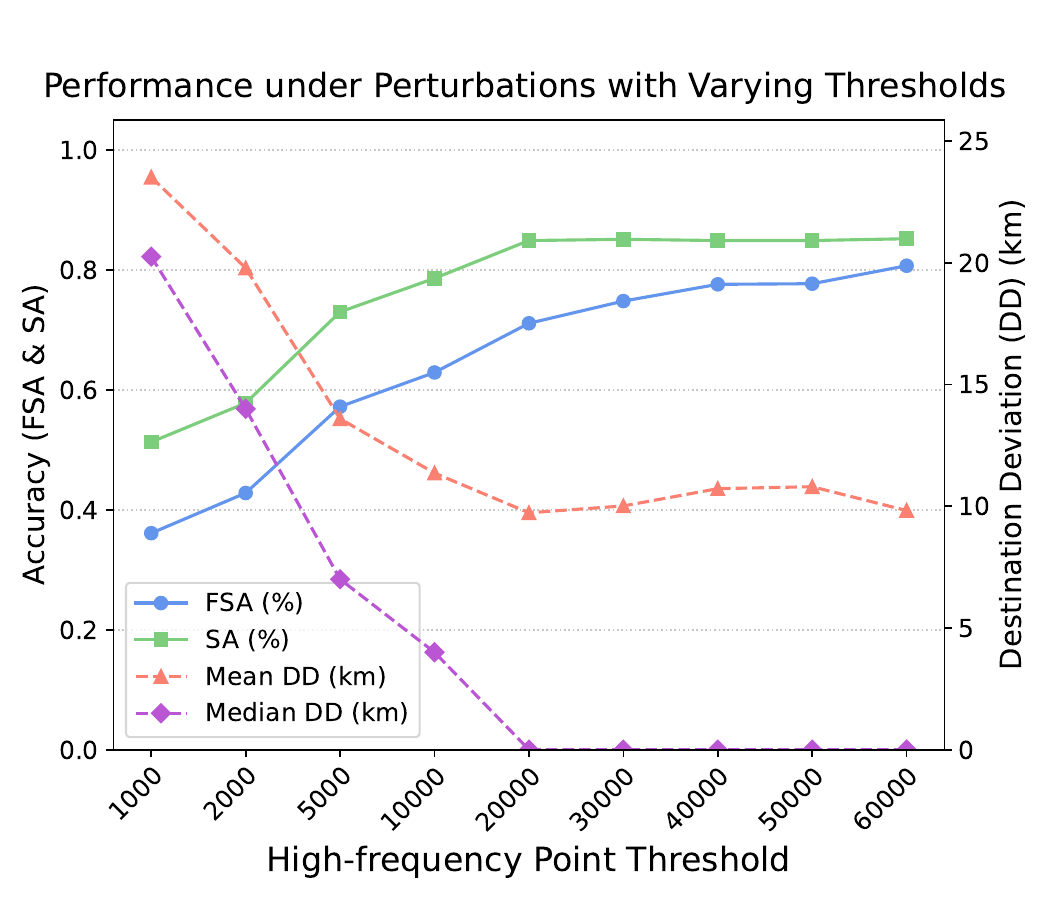}
        \caption{Performance metrics (FSA, SA, Mean/Median DD) versus high-frequency point thresholds. Left y-axis: FSA/SA; Right y-axis: DD (km).}
        \label{fig:train_method_exp8}
    \end{minipage}
\end{figure*}

In addition, we further train the continual pre-trained model \modelone using the sft approach for the shortest path task, and evaluate its robustness against disturbances.
The experimental results are shown in Table~\ref{tab:train_method_exp7} and Figure~\ref{fig:train_method_exp8}.

Experimental results show that Cognition-\textit{ModelTwo} trained via SFT exhibits robustness comparable to that of the CPT-trained counterpart, with both being influenced by the training data distribution—performing better at critical points with larger thresholds. 
Meanwhile, when facing random disturbances, the SFT-trained model reaches destinations closer to the target on average, but demonstrates a significantly lower proportion of selecting valid roads at each step.

\begin{table}[h!]
  \centering
  \small
  \begin{tabular}{ccccc}
    \toprule
    \textbf{Method} & \textbf{FSA (\%)} & \textbf{SA (\%)} & \textbf{DD (km)} \\
    \midrule
    No Pert. & 100.00 & 100.00 & 0.00 \\
    Road Pert. & 8.14 & 52.31 & 12.42 \\
    Distance Pert. & 14.95 & 60.29 & 9.46 \\
    Direction Pert. & 6.01 & 59.53 & 40.89 \\
    \bottomrule
  \end{tabular}
  \caption{Evaluation Results for Different Types of Perturbations Trained via SFT.}
  \label{tab:train_method_exp7}
\end{table}

\subsection{Model Architecture and Scale}

To investigate the impact of \model architecture and scale on spatial cognition, in addition to the Qwen2.5-0.5B used in the main experiments, we further examine the performance of Qwen2.5-1.5B and LLaMA-3.2-1B~\cite{llama3modelcard}.

\paragraph{Spatial Cognition}

The results are shown in Table~\ref{tab:model_exp1}, Table~\ref{tab:model_exp2} and Table~\ref{tab:model_exp4}.

The experiment shows that both larger model sizes and models with other architectures exhibit consistent experimental conclusions.

\begin{table}[t!]
  \centering
  \small
  \begin{tabular}{c cc cc}
    \toprule 

    \multirow{2}{*}{\textbf{Probe Type}} & \multicolumn{2}{c}{\textbf{Distance}} & \multicolumn{2}{c}{\textbf{Azimuth}} \\
    \cmidrule(lr){2-3} \cmidrule(lr){4-5}

     & MAE (km) & R² & MAE (°) & Spearman  \\ 
    \midrule

    Non-linear & 0.85 & 1.00 & 3.49 & 0.98 \\
    Linear & 17.89 & 0.20 & 51.94 & 0.78 \\

    \bottomrule
  \end{tabular}
  \caption{Latent spatial composition evaluation. An MLP predicts distance and azimuth between POI pairs using their concatenated hidden states.}
  \label{tab:linear_exp2}
\end{table}

\begin{table*}[t!]
  \centering
  \small
  \begin{tabular}{c yyy bbb kk}
    \toprule 

    \multirow{2}{*}{\textbf{Probe Type}} & \multicolumn{3}{c}{\textbf{X}} & \multicolumn{3}{c}{\textbf{Y}} & \multicolumn{2}{c}{\textbf{Euclidean Distance}} \\
    \cmidrule(lr){2-4} \cmidrule(lr){5-7} \cmidrule(lr){8-9}

     & \multicolumn{1}{c}{\textbf{MSE}~$\downarrow$} & \multicolumn{1}{c}{\textbf{MAE}~$\downarrow$} & \multicolumn{1}{c}{\textbf{R²}~$\uparrow$} & \multicolumn{1}{c}{\textbf{MSE}~$\downarrow$} & \multicolumn{1}{c}{\textbf{MAE}~$\downarrow$} & \multicolumn{1}{c}{\textbf{R²}~$\uparrow$} & \multicolumn{1}{c}{\textbf{Mean}~$\downarrow$} & \multicolumn{1}{c}{\textbf{Std.}~$\downarrow$} \\ 
    \midrule

    \rowcolor{gray!30}
    \multicolumn{9}{c}{\textit{Absolute Coordinate Probing}} \\
    Non-linear & 1.16 & 0.78 & 1.00 & 0.91 & 0.71 & 1.00 & 1.18 & 0.82 \\
    Linear & 21.18 & 3.61 & 0.97 & 12.70 & 2.75 & 0.99 & 4.99 & 2.99 \\

    \rowcolor{gray!30}
    \multicolumn{9}{c}{\textit{Step-wise Coordinates Probing}} \\
    Non-linear & 6.51 & 1.84 & 0.99 & 6.96 & 1.94 & 0.99 & 3.01 & 2.60 \\
    Linear & 238.65 & 11.97 & 0.68 & 228.98 & 11.73 & 0.69 & 18.68 & 10.86 \\

    \bottomrule
  \end{tabular}
  \caption{Performance of the MLP probe in predicting the absolute coordinates of POIs and dynamic position coordinates at each step of the generated navigation path from the \model’s last hidden states.}
  \label{tab:linear_exp1}
\end{table*}

\begin{table*}[htbp]
  \centering
  \small
  \begin{tabular}{c ccc ccc cc}
    \toprule

    \multirow{2}{*}{\textbf{Model}} & \multicolumn{3}{c}{\textbf{X}} & \multicolumn{3}{c}{\textbf{Y}} & \multicolumn{2}{c}{\textbf{Euclidean Distance}} \\
    \cmidrule(lr){2-4} \cmidrule(lr){5-7} \cmidrule(lr){8-9}

     & MSE~$\downarrow$ & MAE~$\downarrow$ & R²~$\uparrow$ & MSE~$\downarrow$ & MAE~$\downarrow$ & R²~$\uparrow$ & Mean~$\downarrow$ & Std.~$\downarrow$ \\ 
    \midrule

    Qwen2.5-0.5B & 1.16 & 0.78 & 1.00 & 0.91 & 0.71 & 1.00 & 1.18 & 0.82 \\
    Qwen2.5-1.5B & 6.83 & 1.96 & 0.99 & 3.40 & 1.47 & 1.00 & 2.73 & 1.66 \\
    LlaMA-3.2-1B & 5.71 & 1.94 & 0.99 & 6.97 & 1.99 & 0.99 & 3.07 & 1.81 \\

    \bottomrule
  \end{tabular}
  \caption{Performance of the MLP probe in predicting the absolute coordinates of POIs from the \model’s last hidden states under different models.}
  \label{tab:model_exp2}
\end{table*}

\begin{table*}[t!]
    \centering
    \small
    \renewcommand{\arraystretch}{1.3}
    
    \begin{minipage}{0.48\textwidth}
        \centering
        \begin{tabular}{c cc cc}
            \toprule
            \multirow{2}{*}{\textbf{Model}} & \multicolumn{2}{c}{\textbf{Distance}} & \multicolumn{2}{c}{\textbf{Azimuth}} \\
            \cmidrule(lr){2-3} \cmidrule(lr){4-5}
            & MRPE~$\downarrow$ & R²~$\uparrow$ & MRPE~$\downarrow$ & Spearman~$\uparrow$ \\ 
            \midrule
            Qwen2.5-0.5B & 0.11 & 1.00 & 0.79 & 1.00 \\ 
            Qwen2.5-1.5B & 0.28 & 1.00 & 1.30 & 0.99 \\
            LlaMA-3.2-1B & 1.71 & 1.00 & 3.99 & 0.98 \\
            \bottomrule
        \end{tabular}
        \caption{The performance of the model's prediction of distance and azimuth for unseen POI pairs under different models.}
        \label{tab:model_exp1}
    \end{minipage}\hfill%
    
    \begin{minipage}{0.48\textwidth}
        \centering
        \begin{tabular}{c cc cc}
            \toprule
            \multirow{2}{*}{\textbf{Model}} & \multicolumn{2}{c}{\textbf{Distance}} & \multicolumn{2}{c}{\textbf{Azimuth}} \\
            \cmidrule(lr){2-3} \cmidrule(lr){4-5}
            & MAE (km) & R² & MAE (°) & Spearman  \\ 
            \midrule
            Qwen2.5-0.5B & 0.85 & 1.00 & 3.49 & 0.98 \\
            Qwen2.5-1.5B & 1.61 & 1.61 & 5.81 & 0.97 \\
            LlaMA-3.2-1B & 1.18 & 1.00 & 4.32 & 0.96 \\
            \bottomrule
        \end{tabular}
        \caption{Latent spatial composition evaluation. An MLP predicts distance and azimuth between POI pairs using their concatenated hidden states.}
        \label{tab:model_exp4}
    \end{minipage}
\end{table*}

\begin{table*}[t!]
  \centering
  \small
  \begin{tabular}{cccccccc}
    \toprule
     \multirow{2}{*}{\textbf{Model}} & \multicolumn{4}{c}{\textbf{Accuracy}} & \multicolumn{3}{c}{\textbf{Consistency}} \\
    \cmidrule(lr){2-5} \cmidrule(lr){6-8} 
    
     & \multicolumn{1}{c}{\textbf{SPD}~$\downarrow$} & \multicolumn{1}{c}{\textbf{EPD}~$\downarrow$} & \multicolumn{1}{c}{\textbf{VRP} ($\uparrow$\%)} & \multicolumn{1}{c}{\textbf{SPA} ($\uparrow$\%)} & \multicolumn{1}{c}{\textbf{VMR} ($\uparrow$1.0)} & \multicolumn{1}{c}{\textbf{VCS} ($\uparrow$1.0)} & \multicolumn{1}{c}{\textbf{FD} ($\downarrow$0.0)} \\
    \midrule
    LlaMA-3.2-1B & 32.18 & 1.16 & 96.5 & 27.4 & 1.05 & 0.74 & 23.30 \\
    Qwen2.5-1.5B & 0.04 & 0.27 & 97.6 & 89.9 & 1.00 & 1.00 & 0.59 \\
    \bottomrule
  \end{tabular}
  \caption{Performance of different training settings on shortest path prediction between POIs in $P_{\text{heldout}}$.}
  \label{tab:model_navigation}
\end{table*}

\paragraph{Spatial Navigation}

The results are shown in Table~\ref{tab:model_navigation}. The experimental results show that the LLaMA model performs poorly in learning local path information and completing shortest path navigation, with a significant bias in identifying the starting point.

\subsection{Linear vs. Non-linear Probe}

\paragraph{Setup} We use the LinearRegression model from scikit-learn. It relies on a direct mathematical solution to find the best-fit line, and we used its default configuration. For the non-linear probe, we use the same MLP configuration as in the main experiment.

\paragraph{Results} We use \modelone and Base-\modeltwo to compare linear and non-linear probes in several experiments involving probing. The experimental results are shown in the Table~\ref{tab:linear_exp1}, Table~\ref{tab:linear_exp2}.

\paragraph{Conclusion} 
The results in Table~\ref{tab:linear_exp1} demonstrate that a linear probe can map hidden states to actual coordinate values, indicating the presence of linearly accessible coordinate information within the hidden representations of the \model. However, non-linear regression achieves higher prediction accuracy. Furthermore, in the \model trained on shortest-path trajectory data, the performance of the linear probe deteriorates significantly, with the average Euclidean distance increasing from 3.01 to 18.68. This suggests that non-linear probes are better suited for capturing position information in more complex tasks.

The experimental results in Table~\ref{tab:linear_exp2} show that when performing regression to predict distance and azimuth by combining the hidden states of two POIs, the linear probe performs poorly (R² of only 0.20 for distance prediction). 
This suggests that we cannot achieve combined prediction through simple linear regression, which may also be related to how we process the two POI vectors (\eg, concatenation).

\section{Other Statements}
Our use of existing artifacts are consistent with their intended use, and we follow their license and terms.

\end{document}